\documentclass{article}
\usepackage[preprint]{neurips_2025}

\usepackage{tcolorbox}
\usepackage{enumitem}
\usepackage{graphicx}
\usepackage{multirow}
\tcbuselibrary{breakable}

\usepackage{adjustbox}
\usepackage{amsmath}
\usepackage{amssymb}
\usepackage{wrapfig}
\usepackage{multirow}
\usepackage{colortbl}
\usepackage{booktabs}
\usepackage{footmisc}
\usepackage[pagebackref=true,breaklinks=true,colorlinks,bookmarks=false]{hyperref}

\usepackage[dvipsnames]{xcolor}

\def\model{VidEmo}
\def\dataset{Emo-CFG}

\definecolor{blue3}{HTML}{5D8DFD}
\definecolor{green3}{HTML}{88B06D}
\definecolor{orange3}{HTML}{F5A83D}
\definecolor{red3}{HTML}{F5433D}
\definecolor{blue1}{HTML}{AEC6FE}
\definecolor{green1}{HTML}{B3CDA2}
\definecolor{orange1}{HTML}{F9CB8A}
\definecolor{red1}{HTML}{FBA09D}
\definecolor{lightgray}{gray}{1.0}
\definecolor{cambriangray}{gray}{0.9}
\definecolor{LightCyan}{rgb}{0.88,1,1}

\definecolor{LightCyan}{rgb}{0.88,1,1}
\definecolor{ProcessBlue}{rgb}{0,0.4,0.8}
\hypersetup{
    colorlinks=true,
    linkcolor=ProcessBlue,
    citecolor=ProcessBlue
}
\newcommand\blfootnote[1]{
    \begingroup
    \renewcommand\thefootnote{}\footnote{#1}
    \addtocounter{footnote}{-1}
    \endgroup
}

\definecolor{cvprblue}{rgb}{0.21,0.49,0.74}
\usepackage{hyperref}

\title{\model: Affective-Tree Reasoning for Emotion-Centric Video Foundation Models}

\author{
\textbf{Zhicheng Zhang$^{1*}$, Weicheng Wang$^{1*}$, Yongjie Zhu$^{3\dag}$}\\
\textbf{Wenyu Qin$^3$, Pengfei Wan$^3$, Di Zhang$^3$, Jufeng Yang$^{124\ddag}$}\\
{\small $^1$ Nankai University}\ 
{\small $^2$ Pengcheng Laboratory}\ 
{\small $^3$ Kuaishou Technology}\\
{\small $^4$ Nankai International Advanced Research Institute (SHENZHEN·FUTIAN)}\\
{\small
\texttt{gloryzzc6@sina.com},
\texttt{2120230639@mail.nankai.edu.cn}}\\
{\small
\texttt{\{zhuyongjie,qinwenyu,wanpengfei\}@kuaishou.com},
\texttt{yangjufeng@nankai.edu.cn}
}\\ 
\textbf{\textit{\href{https://zzcheng.top/VidEmo}{https://zzcheng.top/VidEmo}}}
}

\begin{document}
\maketitle
\blfootnote{$*$: Equal Contribution. $\dag$: Project Leader.  $\ddag$: Corresponding Author.}
\begin{figure}[h]
  \centering
  \vspace{-35pt}
  \includegraphics[width=1.0\linewidth]{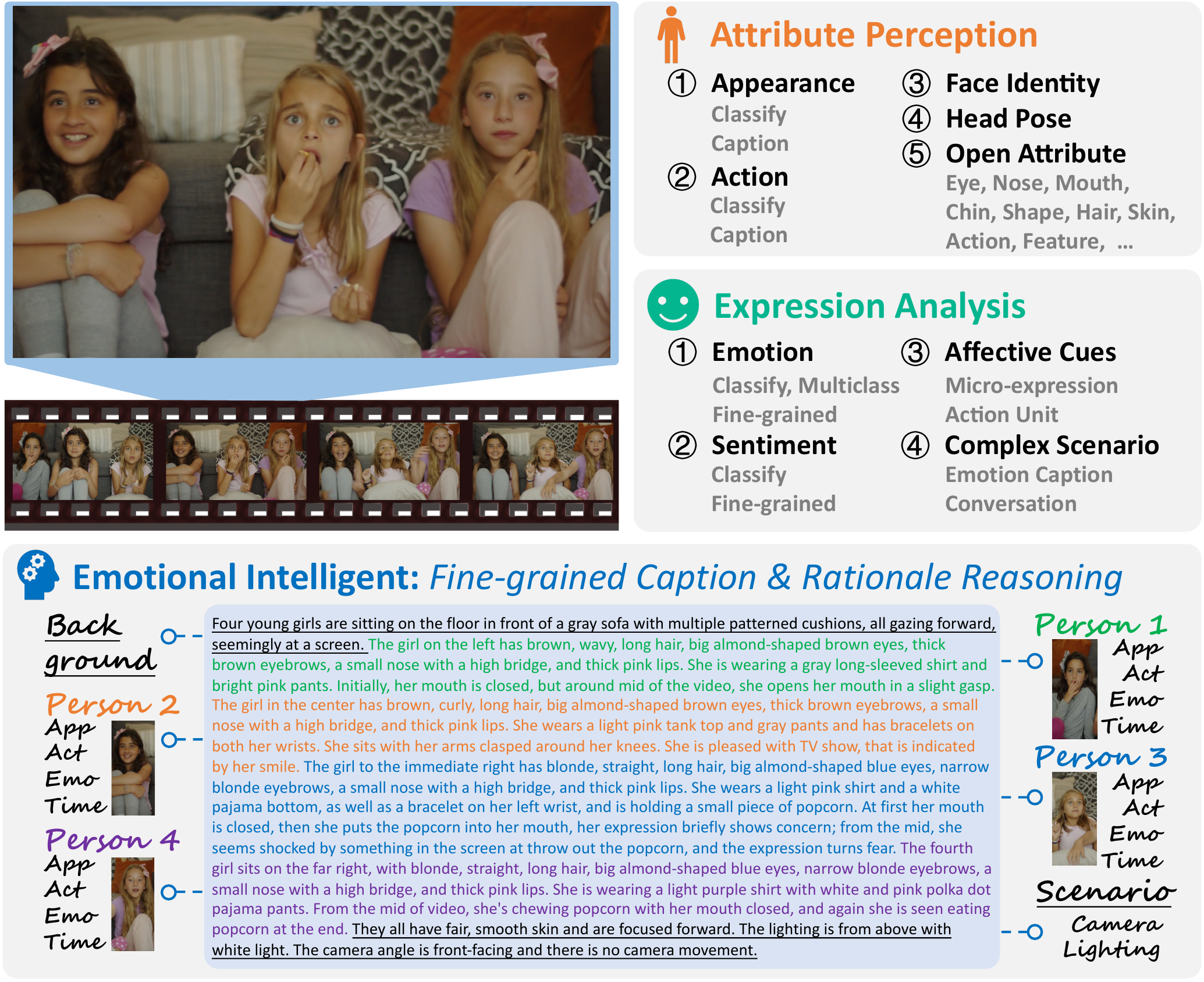}
  \vspace{-20pt}
  \caption{\textbf{Selected examples of inputs and outputs obtained from~\model}. Apart from providing toolkits for basic attribute perception and expression analysis (top), \model~extends the cognitive capacity and is able to generate fine-grained emotional captions with explainable rationale (bottom).}
\vspace{-15pt}
\label{fig:motivation}
\end{figure}

\begin{abstract}
Understanding and predicting emotion from videos has gathered significant attention in recent studies, driven by advancements in video large language models (VideoLLMs).
While advanced methods have made progress in video emotion analysis, the intrinsic nature of emotions poses significant challenges.
Emotions are characterized by dynamic and cues-dependent properties, making it difficult to understand complex and evolving emotional states with reasonable rationale.
To tackle these challenges, we propose a novel affective cues-guided reasoning framework that unifies fundamental attribute perception, expression analysis, and high-level emotional understanding in a stage-wise manner.
At the core of our approach is a family of video emotion foundation models (\model), specifically designed for emotion reasoning and instruction-following.
These models undergo a two-stage tuning process: first, curriculum emotion learning for injecting emotion knowledge, followed by affective-tree reinforcement learning for emotion reasoning.
Moreover, we establish a foundational data infrastructure and introduce a emotion-centric fine-grained dataset (\dataset) consisting of 2.1M diverse instruction-based samples.
\dataset~includes explainable emotional question-answering, fine-grained captions, and associated rationales, providing essential resources for advancing emotion understanding tasks.
Experimental results demonstrate that our approach achieves competitive performance, setting a new milestone across 15 face perception tasks.
\end{abstract}

\section{Introduction}

Understanding and predicting human emotions from dynamic videos is an increasingly vital challenge in computer vision, with far-reaching applications in human-computer interaction, surveillance, and healthcare~\cite{grgic2011scface,preece1994human,sardar2023secure}.
Despite the success of advanced methods~\cite{lian2024affectgpt,xie2024emovit,xing2024emo}, particularly in classifying basic emotional expressions, the ability to predict about complex, evolving emotional states with reasonable rationale remains limited.
This is largely due to the dynamic and context-dependent nature of emotions~\cite{zhao2021emotion,lian2024ov}, which require models capable of providing both high-level emotional intelligence and rational, explainable outputs~\cite{venkatraman2020logic,lian2023explainable}.
Recently, the emergence of VideoLLMs~\cite{chen2024expanding,qwen2025qwen,ye2024mplug,zhang2025videollama} has provided a promising baseline as a pathway.
However, these foundational models often struggle with high-level emotional understanding, as they lack the ability to effectively combine basic facial attributes into representations of complex emotion.
Even the cutting-edge milestone, Gemini 2.0~\cite{team2024gemini}, achieves only an accuracy of 26.3\% in fine-grained sentiment analysis, highlighting the gap in performance and the need for further innovation in this domain.

\begin{figure}[!bp]
  \centering
  \vspace{-10pt}
  \includegraphics[width=1.0\linewidth]{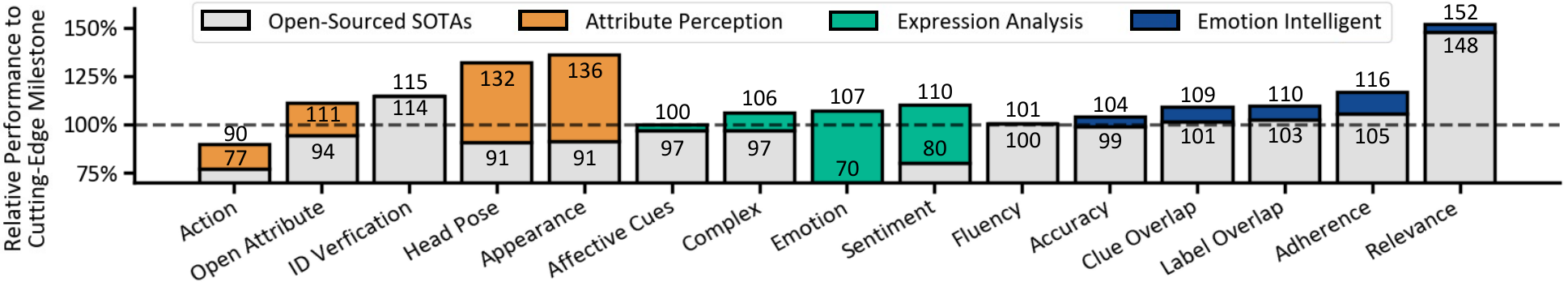}
  \vspace{-22pt}
  \caption{\textbf{Results Overview.} Our best model, \textit{VidEmo}-T1, shows superior performance across 15 face perception tasks, surpassing advanced milestone (\textit{Gemini 2.0: 5\textsuperscript{th} Feb, 2025}) on 14 of 15 tasks.}
    \vspace{-10pt}
  \label{fig:perf}
\end{figure}

To address these challenges, we introduce \model, a novel affective cues-guided reasoning framework based on tree-structure that integrates three core components: fundamental attribute perception, expression analysis, and high-level emotional understanding (see Fig.~\ref{fig:motivation}).
Across 15 face perception tasks, \model~outperforms all existing open-source VideoLLMs, surpassing the previous state-of-the-art benchmark, \textit{i.e.}, Gemini 2.0, as shown in Fig.\ref{fig:perf}.
To achieve this, Our \model~draws inspiration from recent work on reasoning models (R1), which excel at providing explainable rationales~\cite{guo2025deepseek,openai2025o3-mini,qwen2024qvq,xiang2025towards}.
These models solve complex tasks by incorporating a thinking process alongside the model’s operation.
Our finding demonstrates that this same reasoning process can be applied to high-level emotion understanding by introducing stage-wise thinking, structured around attribute perception~\cite{zheng2020survey,cai2023marlin}, expression analysis~\cite{ben2021video,sun2023mae}, and emotion understanding~\cite{zhang2023weakly,wang2015video}.
To be specific, we equip \model~with curriculum emotion learning and affective-tree reasoning, which inject emotion reasoning pathways during both the pre-train and post-train stages, respectively.
In the pre-train stage, curriculum emotion learning progressively tunes the model from basic facial attributes to more complex emotional states.
In the post-train stage, affective-tree reasoning helps the model refine its emotional understanding by using a hierarchical structure, ensuring that emotional responses are both accurate and interpretable.
This two-stage process enables \model~to effectively analyze and reason about emotions in dynamic video data.

To support our approach, we construct an emotion-centric fine-grained dataset, \dataset, specifically designed to serve as the foundational data infrastructure for emotion understanding.
\dataset~is a large-scale dataset consisting of 2.1 million diverse, characterized by emotion-centric labels, rigorous data verification, and high diversity, ensuring comprehensive and reliable annotations across a wide range of emotional contexts.
By offering rich annotations and a wide variety of emotional contexts, Emo-CFG empowers \model~to effectively learn fine-grained emotion understanding from the emotion reasoning pathway.

Our contributions are two-fold:
(1) We propose \textbf{\model}, a novel affective cues-guided reasoning framework that combines curriculum emotion learning and affective-tree reasoning, enabling fine-grained and interpretable emotion understanding from dynamic video data.
Experimental results show that \model~achieves over a 16.3\% and 14.2\% improvement compared to existing open-source VideoLLMs across 15 facial perception tasks on 1-3B and 7-8B scales.
(2) We present \textbf{\dataset}, a large-scale, emotion-centric dataset comprising 2.1M diverse samples with detailed annotations across attributes, expressions, and emotions, serving as a comprehensive data infrastructure for advancing emotion-centric video analysis.

\section{Related Work}

\textbf{Facial Video Analysis.}
Face video analysis is a long standing problem towards high-level human understanding which involves various tasks, including attribute perception~\cite{zheng2020survey,cai2023marlin}, expression analysis~\cite{ben2021video,sun2023mae}, and emotion understanding~\cite{zhang2023weakly,wang2015video,li2022multi}.
Various face perception models leverages strong backbone power for constructing multi-task framework~\cite{sun2024face}.
Going forward to high-level emotion understanding~\cite{zhang2024masked,zhao2021emotion,zhang2022temporal}, recent methods embrace MLLM~\cite{chen2024far,li2024facial,lian2025affectgpt,zhang2025moda} for their strong zero-shot perception capacity~\cite{sun2024face,zhang-etal-2024-visual,Etesam2024ContextualER}. 
EmotionLLaMA~\cite{cheng2024emotionllama} introduces an emotion dataset including 28K coarse-grained and 4K fine-grained annotated datasets.
OmniEmotion~\cite{yang2025omni} proposes to explicitly integrate facial and audio modeling for emotion recognition. 
However, existing approaches are often constrained to a limited set of emotion categories or rely on static attribution perception.
To advance cognitive human emotion understanding, we propose a fine-grained emotion-centric model empowered by dynamic attribution perception and emotion reasoning.

\textbf{Reasoning Model in MLLM.}
With the blossom of a series of recent models such as DeepSeek-R1 and OpenAI o-series~\cite{openai2025o3-mini,guo2025deepseek}, various works probe into integrating MLLMs with reasoning capacity~\cite{ahn-etal-2024-large}.
Multimodal chain-of-thought (MCoT) prompting~\cite{li2025imagine,thawakar2025llamav,liu2024ocrbench} offers a step-by-step reasoning trajectory when MLLM faces hard questions including detail grounding~\cite{wu2024v,liu2024mminstruct}, agent planing~\cite{li2025imagine}, etc.
Specifically, MCoT aims to tackle the question through several solving steps and a reasoning chain, enabling the generation of more effective results for complex problems step-by-step~\cite{xiang2025towards,qwen2024qvq,zhang2024improve}.
For instance, LLaVA-CoT~\cite{xu2024llava} prompts MLLMs reasoning steps into the summary, caption, reasoning, and conclusion stages and proposes a stage-level beam search strategy to further enhance reasoning capacity.
In this paper, we propose affective cues-based rationale tree as intermediate bridge to meet the gap between abstract emotion and basic attribute.

\section{\model: Video Emotion Foundation Models}
\label{sec:method}
\begin{figure}[!tbp]
    \centering
    \vspace{-20pt}
    \includegraphics[width=1.0\linewidth]{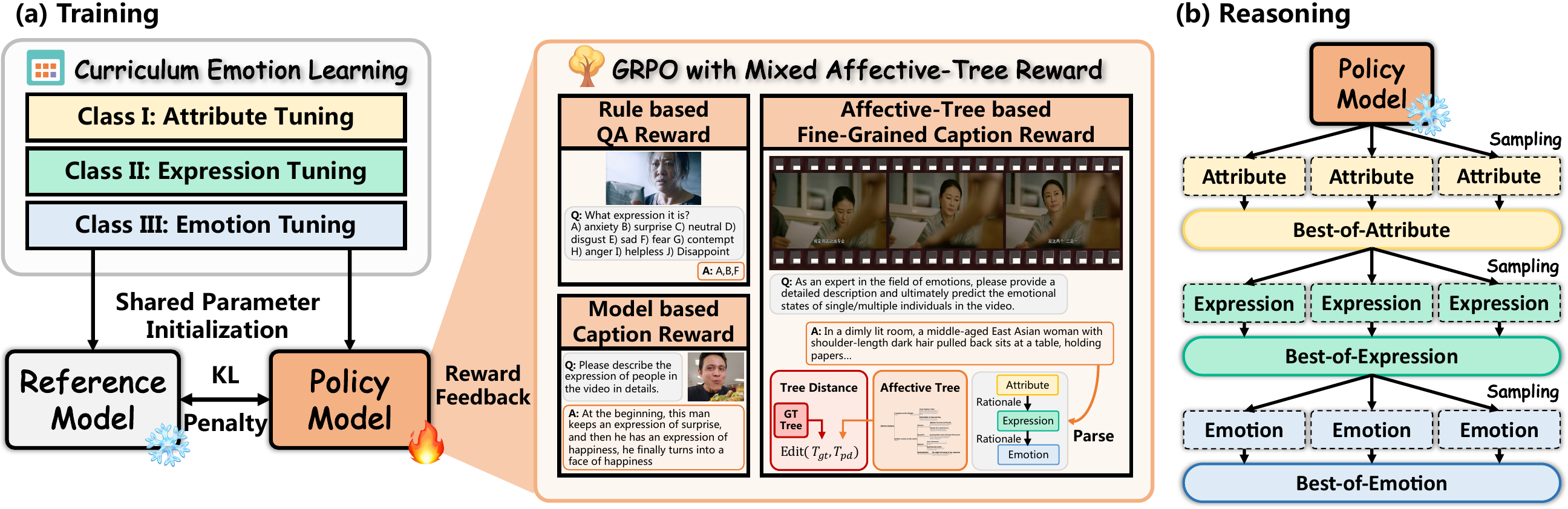}
    \vspace{-15pt}
    \caption{\textbf{Pipeline of \model.}
    (a) Training: The model is trained using curriculum emotion learning, divided into three stages: attribute, expression, and emotion tuning. A reference model provides initial parameters, and a policy model is trained with reward feedback.
    (b) Reasoning: The policy model performs hierarchical reasoning by sampling from the best attributes, expressions, and emotions to generate the final emotional output.
    }
    \vspace{-20pt}
  \label{fig:pipeline}
\end{figure}

To develop a family of emotion-centric video foundation models, we propose a comprehensive set of toolkits designed for the pre-training, post-training, and reasoning stages, as illustrated in Fig.~\ref{fig:pipeline}.
Through a structured pre-training process, emotion knowledge is injected, followed by post-training to enhance the model’s reasoning capabilities.
Finally, the reasoning stage allows the model to effectively generate emotional outputs, leveraging learned attributes, expressions, and emotions.

\subsection{Pre-training: Curriculum Emotion Learning}
\begin{figure}[!tbp]
    \centering
    \vspace{-25pt}
    \includegraphics[width=\linewidth]{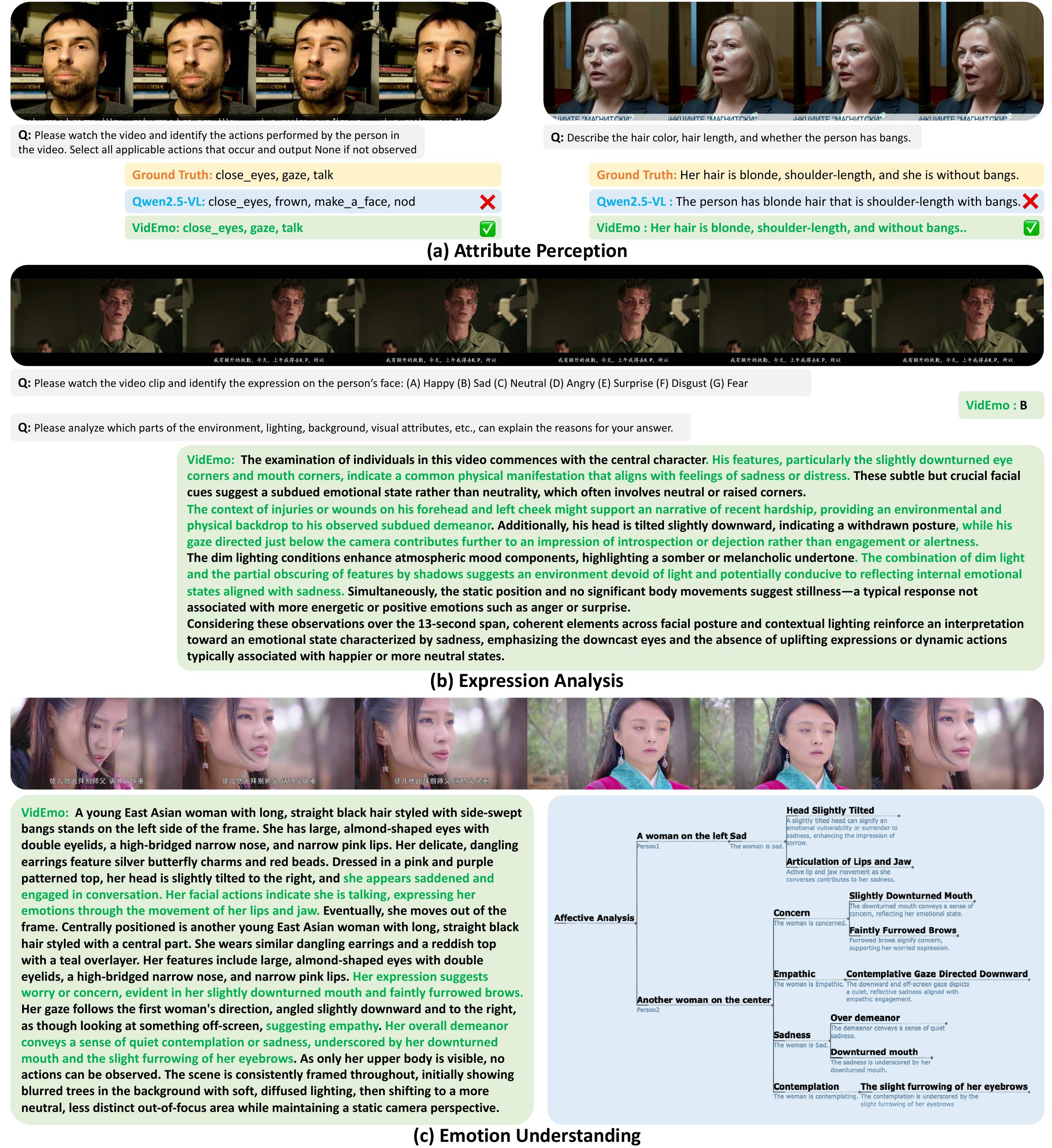}
    \vspace{-10pt}
    \caption{\textbf{Visualization} on attribute perception, expression analysis, and emotion understanding.}
    \vspace{-15pt}
    \label{fig:Visualization}
\end{figure}

To inject emotion knowledge into the foundation model, we employ curriculum emotion learning to progressively tuning our base model.
The training is structured into three stages: I) Attribute Tuning, II) Expression Tuning, and III) Emotion Tuning.
The pre-training focuses on curating data that balances the difficulty of emotion tasks while addressing perplexity.
At each stage, we carefully curate the data to ensure that the emotion-related tasks gradually increase in complexity.
By starting with simpler attributes and progressively moving towards more complex expressions and emotions, we ensure that the model builds a strong foundational understanding of emotion, which facilitates smoother emotion knowledge injection throughout the process.
Figure~\ref{fig:Visualization} presents the visualization results of our model across three key aspects: Attribute Perception, Expression Analysis, and Emotion Understanding.

\textbf{Attribute Perception}: The model accurately identifies facial attributes, such as hair color, length, and presence of bangs, with the ground truth comparison clearly shown for validation. For instance, the model correctly identifies a person’s hair as blonde and shoulder-length, while also distinguishing the presence or absence of bangs.

\textbf{Expression Analysis}: The model analyzes subtle facial expressions, identifying features such as downward-tilted eyes and posture. These features, as seen in the second part of the figure, provide insight into the emotional states of the person, such as sadness or introspection, based on facial and contextual cues, like lighting and body movements.

\textbf{Emotion Understanding}: By combining the insights from facial features and contextual cues, the model provides a detailed interpretation of the emotional state. For example, in the final part of the figure, the model identifies a contemplative emotion, indicated by the subject’s slightly tilted head, furrowed brows, and subtle eye movements.

\subsection{Post-training: GRPO via Mixed Affective-Tree Reward}

Building on emotion-knowledge-injected base models, we proceed to post-training to explore the emotional reasoning pathway.
Recent reinforcement learning (RL) techniques~\cite{xiang2025towards} have demonstrated strong capabilities in reasoning, and GRPO~\cite{guo2025deepseek} has garnered significant attention due to its simplicity and effectiveness.
This makes GRPO an ideal starting point for our work.

Formally, let $q$ be a query, GRPO samples a group of outputs ${\{o_i\}}_{i=1}^G$ with the number of $G$ from the old policy model $\pi_{\theta_{\text{old}}}$, and train a policy model by maximizing the following objective:

\resizebox{\textwidth}{!}{
$
\displaystyle
\mathcal{J}_{\text{GRPO}}(\theta) = 
\mathbb{E}_{q, \{o_i\} \sim \pi_{\theta_{\text{old}}}} 
\left[ 
\frac{1}{G} \sum_{i=1}^{G} \frac{1}{|o_i|} \sum_{t=1}^{|o_i|}
\min \left( r_t(\theta) \hat{A}_{i,t}, \,
\text{clip}(r_t(\theta), 1 - \epsilon, 1 + \epsilon) \hat{A}_{i,t} \right)
\right]
- \beta D_{\text{KL}}(\pi_\theta \,\|\, \pi_{\text{ref}}),
$
}
where $\hat{A}_{i,t}$ is the advantage based on relative rewards in group, $\epsilon$ and $\beta$ are coefficient of KL penalty and clip threshold, and \( \pi_{\theta}, \pi_{\theta_{\text{old}}}, \pi_{\text{ref}} \) are current, old, and reference policy models, respectively.

\textbf{Rule based QA Reward.}
The model is evaluated on its ability to respond to emotion-related queries using predefined rules of Acc and F1 score.
The evaluation tasks include classification (single-label, multi-label), fine-grained classification, micro-expression detection, and action unit (AU) detection.

\textbf{Model based Caption Reward.}
For the short caption of action, appearance, and emotion, we use a generative reward model to score the quality of captions generated by the model.

\textbf{Affective-Tree based Fine-Grained Caption Reward.}
To assess the model’s capacity for structured emotional reasoning, we introduce a reward mechanism based on a hierarchical affective tree constructed from fine-grained captions.
Given a generated caption $\hat{o}$, we first parse it into a set of aspect–item pairs at three semantic levels: attribute ($\mathcal{A}$), expression ($\mathcal{E}$), and emotion ($\mathcal{M}$).
These elements are organized into a three-level affective tree $T_{\text{pred}}$, where each node represents an extracted item and directed edges encode rationale-based dependencies:
\begin{equation}
\mathcal{A} \xrightarrow{\text{rationale}} \mathcal{E} \xrightarrow{\text{rationale}} \mathcal{M}.
\end{equation}

We compare the predicted tree $T_{\text{pred}}$ with a ground-truth tree $T_{\text{gt}}$, parsed from human-annotated captions, using the tree edit distance~\cite{zhang1989simple} $\text{Edit}(T_{\text{gt}}, T_{\text{pred}})$, which quantifies the minimal number of edit operations (insertions, deletions, substitutions) required to transform one tree into the other.

The final reward $R$ is computed using an exponential decay over the tree distance:
\begin{equation}
R = \exp\left(-\lambda \cdot \text{Edit}(T_{\text{gt}}, T_{\text{pred}})\right),
\end{equation}

where $\lambda > 0$ is a scaling factor controlling the reward sensitivity to tree differences. This formulation encourages the model to generate captions that are not only accurate in content but also structurally explainable, aligning with human reasoning patterns over emotional understanding.

\subsection{Inference: Reasoning for High-level Emotion Understanding}
Our~\model~facilitates stage-wise training can be smoothly integrated with search-based reasoning strategy.
Specifically, we adopt a hierarchical, search-based reasoning approach that decomposes emotional understanding into three levels: attribute perception, expression analysis, and emotion inference.
At each level, the policy model samples multiple candidate outputs and selects the best one via a reward-guided scoring mechanism, forming a bottom-up reasoning trajectory.
It is notice that we disable ER when comparing with other SOTA methods, for a fair comparison setting with only one model response are sampled.

\section{\dataset: Emotion-Centric Fine-Grained Video Dataset}
\label{sec:dataset}
\begin{figure*}[!tbp]
  \centering
  \vspace{-20pt}
  \includegraphics[width=1.0\linewidth]{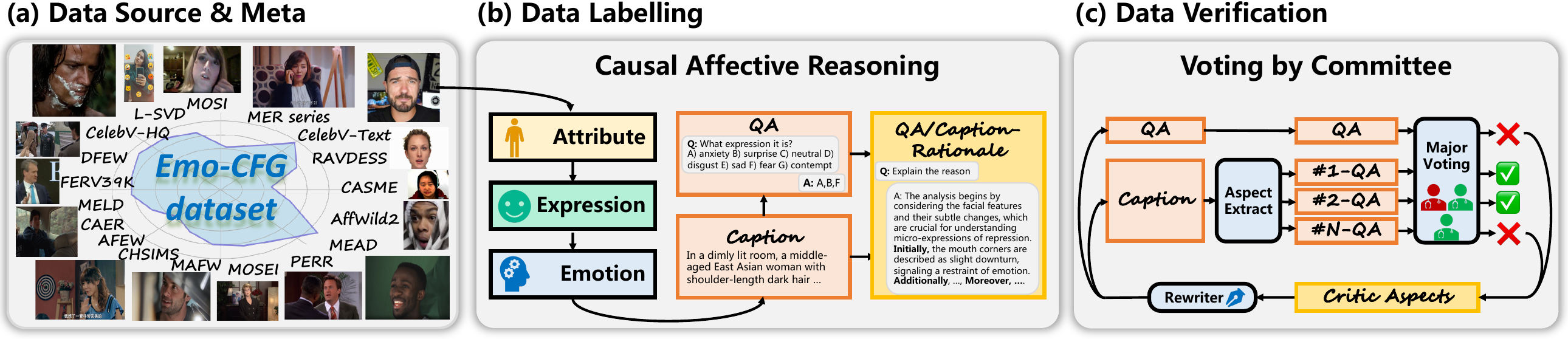}
  \vspace{-20pt}
  \caption{\textbf{Data Curation Pipeline of the \dataset~dataset.}
    (a) The source of data from 17 datasets.
    (b) The illustration of data labeling steps.
    (c) The illustration of data verification loop.
  }
  \vspace{-10pt}
  \label{fig:data_source}
\end{figure*}

\begin{figure*}[!tbp]
  \centering
  \vspace{-20pt}
  \includegraphics[width=1.0\linewidth]{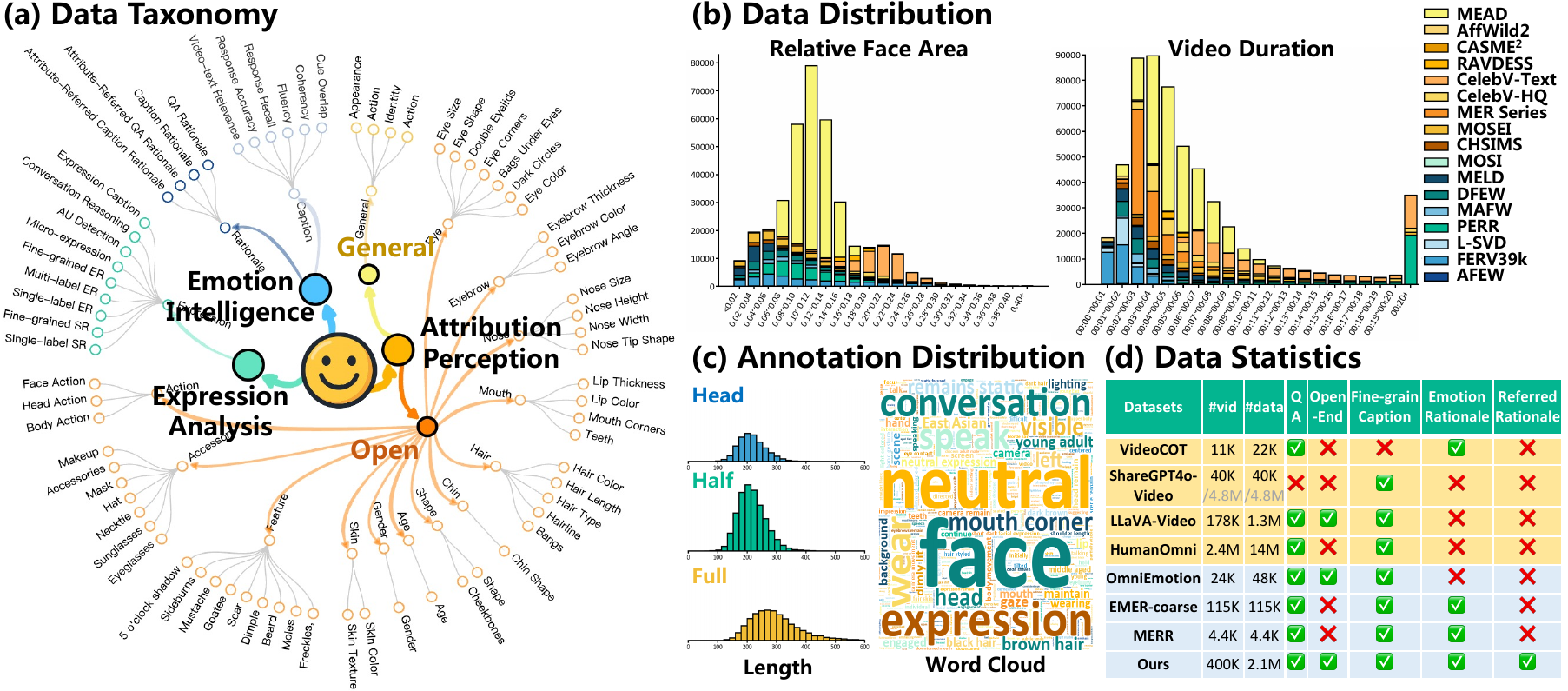}
  \vspace{-20pt}
  \caption{\textbf{Data Statistics of our \dataset~dataset.}
    (a) The data taxonomy from three types of face perception tasks.
    (b) The temporal and spatial distribution of video data.
    (c) The data label distribution and examples.
    (d) The comparison with other emotion and video datasets.
    }
  \label{fig:data_stat}
\end{figure*}

The \dataset~dataset is designed to advance the understanding of emotional dynamics in the video. 
Motivated by the need for high-quality, emotion-centric data to train emotion reasoning models, \dataset~addresses key challenges of diverse emotional contexts, reliable annotations, and rigorous verification.
We illustrate the data curation pipeline and statistics of \dataset~in Fig.~\ref{fig:data_source}\&Fig.~\ref{fig:data_stat}.

\textbf{Data Source \& Meta Information.} 
Our data collecting starts from high-quality video datasets.
The data source include 17 datasets from head, avatar, and full-body avatar.
By utilizing multiple data types, we ensure a holistic perspective to understanding the nuances of visual and emotional data. 
Further, we maintain the meta information of each video, including the face bounding box offset, video duration, video resolution, video fps.

\textbf{$\mathcal{C}{aption}$  \& $\mathcal{QA}$ Instruction Data Labeling.}
We utilize two primary data sources for labeling: large-scale, unlabelled datasets for broad coverage, and small-scale, fully labeled datasets for precision.
For the labeled datasets, instruction pairs are generated using GPT-4o, which creates multiple templates, including multiple-choice questions, open-ended questions, and short captions.
For the unlabelled datasets, we apply a causal affective reasoning strategy to generate labels in a sequential, stage-by-stage manner.
Specifically, given a video, we first leverage the state-of-the-art Gemini 2.0 model, prompting it to generate fine-grained $\mathcal{C}aption$ data, focusing on attributes, expressions, and emotions in sequence.
Subsequently, $\mathcal{QA}$ pairs are generated using GPT-4o, tailored to different aspects of the video.
By combining these attribute and expression labels, the underlying emotion is accurately inferred, enabling a detailed and nuanced understanding of emotional states.

\textbf{$\mathcal{C}aption-\mathcal{R}$  \& $\mathcal{QA-R}$ Rationale Data Labeling.}
Building upon the instruction data, we further explore the relationship between low-level attributes and high-level emotions.
We prompt the advanced VideoLLM to conduct self-reflection on the rationale behind the emotional cues $\mathcal{Q} \stackrel{\mathcal{R}}{\longrightarrow} \mathcal{A}$ and $\mathcal{Q} \stackrel{\mathcal{R}}{\longrightarrow} \mathcal{C}aption$.
This step not only enhances the model’s interpretability by offering insights behind emotional expressions, but also serves as a crucial stage for enabling reasoning capacity.

\textbf{$\mathcal{C}ritic$ Data Verification by Committee Voting.}
To address the inherent ambiguities in emotional data, which arise from its subjective nature, we implement a committee voting-based verification strategy.
We use three heterogeneous VideoLLMs as a committee to verify the correctness of the data and output $\mathcal{C}ritic$ items, including incorrect answers and suggested corrections.
Verified data is retained, while data that does not pass verification is sent back for rewriting based on the suggested corrections.
Additionally, we extract different aspects of the caption data and separate them into multiple QA pairs to ensure alignment with the QA process.

\begin{table*}[!tbp]
\centering
\tabcolsep=2.8pt
\vspace{-10pt}
\caption{\textbf{Comparison with 18 leading VideoLLMs on 14 face attribute perception tasks of~\dataset}, including 6 closed-set attribute perception tasks and 12 open attribute perception tasks. Cls: classification, Cap: caption, ID: identity verification, Pose: head pose estimation, AVG: average.}
\begin{adjustbox}{width=\textwidth}
\begin{tabular}{p{3.05cm}p{.4cm}<{\centering}p{.75cm}<{\centering}p{.75cm}<{\centering}p{.61cm}<{\centering}p{.61cm}<{\centering}p{.61cm}<{\centering}p{.61cm}<{\centering}p{.61cm}<{\centering}p{.65cm}<{\centering}p{.65cm}<{\centering}p{.65cm}<{\centering}p{.65cm}<{\centering}p{.65cm}<{\centering}p{.65cm}<{\centering}p{.65cm}<{\centering}p{.65cm}<{\centering}p{.65cm}<{\centering}p{.65cm}<{\centering}p{.65cm}<{\centering}p{.65cm}<{\centering}p{.7cm}<{\centering}}
\toprule
\multirow{2}{*}{Model} &\multirow{2}{*}{Size}&\multicolumn{2}{c}{Appearance}&\multicolumn{2}{c}{Action}&ID&Head&\multirow{2}{*}{AVG}&\multicolumn{12}{c}{Open Attribute}&\multirow{2}{*}{AVG} \\
\cmidrule(lr){3-4}
\cmidrule(lr){5-6}
\cmidrule(lr){7-7}
\cmidrule(lr){8-8}
\cmidrule(lr){10-21}
&  &Cls&Cap&Cls&Cap&Veri&Pose& &Eye&Mout.&Nose&Hair&Chin&Shap.&Feat.&Acce.&Age&Gend.&Skin&Act. &\\
\midrule
\rowcolor{gray!10}
\multicolumn{4}{l}{\textit{Closed MLLM API}}
            &  &  &  &  &  &  &  &  &  &  &  &  &  &  &  &  &  &   \\
Gemini 2.0~\cite{team2024gemini}
&-&42.2&56.4&41.4&62.2&86.5&25.8&52.4&72.1&72.9&87.3&87.3&64.1&61.9&72.1&80.0&78.7&94.3&81.5&61.6&76.2\\
Claude 3.5 Sonnet~\cite{anthropic2024claude}
&-&39.1&58.8&35.7&61.1&63.6&22.0&46.7&70.8&67.6&54.4&77.8&68.9&56.1&54.7&77.3&79.5&93.3&48.9&61.4&67.6\\
Qwen-VL-MAX~\cite{qwen2023qwen}
&-&41.1&54.3&32.5&60.0&89.7&34.5&52.0&71.9&69.6&79.4&84.4&64.0&71.5&64.4&71.7&78.9&93.3&74.0&60.9&73.7\\
GPT-4o~\cite{OpenAI2024}
&-&29.3&51.5&05.1&40.7&79.0&27.6&38.8&51.5&45.7&62.9&77.1&48.5&46.4&54.5&78.1&68.5&86.3&75.5&52.8&62.3\ \\
GPT-4o mini~\cite{OpenAI2024}
&-  &20.5&55.8&04.1&54.0&69.3&27.8&38.6&43.2&44.5&52.0&52.4&31.6&43.5&40.1&45.0&41.0&69.6&45.9&30.7&44.9  \\

\midrule
\rowcolor{gray!10}
\multicolumn{4}{l}
{\textit{Open-sourced 1-3B Video MLLM}}         &  &  &  &  &  &  &  &  &  &  &  &  &  &  &  &  &  &   \\
LLaVA-OV~\cite{li2024llava}
&1B&06.3&34.9&00.2&47.6&50.3&14.4&25.6&41.0&49.7&36.0&50.9&46.7&32.0&39.0&48.1&29.8&87.6&20.1&61.9&45.2\\
InternVL2.5~\cite{chen2024expanding}
&2B&17.7&46.2&13.4&47.1&04.7&17.3&24.4&53.2&50.6&57.5&70.3&43.5&38.6&42.8&54.5&52.1&80.7&59.8&51.6&54.6\\
VideoLLaMA3~\cite{zhang2025videollama}
&2B&00.3&36.8&05.2&48.5&89.2&20.6&33.4&55.9&51.5&52.0&73.7&45.2&36.7&47.1&52.7&55.0&85.4&59.3&52.1&55.5\\
mPLUG-Owl3~\cite{ye2024mplug}
&2B&16.0&45.4&13.8&\textbf{52.3}&76.4&07.7&35.3&54.4&61.4&55.4&71.4&43.7&45.4&52.3&60.8&39.8&91.6&58.7&48.5&56.9\\
Qwen2.5-VL~\cite{qwen2025qwen}
&3B&43.6&41.1&30.2&49.9&95.7&15.5&46.0&64.2&54.3&51.3&72.8&29.1&40.8&52.1&58.9&70.6&93.5&76.2&\textbf{62.6}&60.5\\
\midrule
\rowcolor{LightCyan}
VidEmo-Base&3B
&\textbf{57.0}&\textbf{67.9}&\textbf{37.7}&47.9&\textbf{100}&\textbf{90.7}&\textbf{66.9}&\textbf{84.9}&\textbf{82.7}&\textbf{94.0}&\textbf{85.2}&\textbf{75.9}&\textbf{80.8}&\textbf{78.0}&\textbf{83.4}&\textbf{84.0}&\textbf{95.0}&\textbf{88.8}&61.1&\textbf{82.8} \\
\midrule
\rowcolor{gray!10}
\multicolumn{4}{l}
{\textit{Open-sourced 7B+ Video MLLM}}          &  &  &  &  &  &  &  &  &  &  &  &  &  &  &  &  &  &   \\
ShareGPT4Video~\cite{chen2024sharegpt4video}
&8B&10.3&38.7&13.7&51.6&03.0&17.1&22.4&53.9&54.7&37.7&74.8&13.1&28.6&46.7&45.1&45.2&51.6&57.9&53.1&46.9\\
InternVL2.5~\cite{chen2024expanding}
&8B&36.9&38.7&17.4&49.8&61.2&15.5&36.6&56.3&59.0&55.0&72.2&52.4&36.4&52.7&61.5&60.0&76.9&57.6&59.3&58.3\\
LLaVA-N-Video~\cite{liu2024llavanext}
&7B&16.9&34.6&20.5&49.2&51.7&05.8&29.8&42.6&43.0&40.4&67.5&18.3&49.9&44.2&52.3&16.7&84.4&58.2&48.6&47.2\\
LLaVA-OV~\cite{li2024llava}
&7B&05.6&37.3&12.2&46.6&97.2&19.8&36.4&53.0&47.6&50.4&64.0&35.3&32.2&49.9&55.8&72.9&94.6&47.5&59.0&55.2\\
VideoLLaMA3~\cite{zhang2025videollama}
&7B&28.3&33.5&15.8&48.7&89.2&16.4&38.6&54.4&56.7&55.5&71.9&40.5&36.6&50.6&61.3&60.4&84.7&65.6&64.8&58.6\\
LLaVA-Video~\cite{zhang2024video}
&7B&14.5&38.2&14.1&46.0&88.7&20.3&37.0&65.0&57.0&66.2&72.6&21.3&29.4&59.3&68.3&79.4&93.0&62.5&63.8&61.5\\
mPLUG-Owl3~\cite{ye2024mplug}
&7B&34.3&41.6&21.3&{55.1}&66.4&21.1&40.0&55.0&56.2&48.0&70.5&39.9&48.1&50.5&58.8&61.8&89.7&62.0&60.8&58.4\\
Qwen2.5-VL~\cite{qwen2025qwen}
&7B&44.7&45.2&21.0&52.3&99.7&22.6&47.6&68.6&70.5&83.2&74.7&66.4&55.6&60.8&73.8&77.2&94.0&76.2&64.0&72.1\\
\midrule
\rowcolor{LightCyan}
\model-Base
&7B&{60.3}&{72.9}&{38.4}&{55.1}&\textbf{99.8}&{93.4}&{69.2}&{86.4}&{85.5}&{95.1}&{85.1}&{77.3}&{81.6}&{78.7}&{85.0}&{85.6}&{95.0}&{89.5}&{71.7}&{84.7}\\
\rowcolor{LightCyan}
\model-T1
&7B&\textbf{64.8}&\textbf{73.1}&\textbf{41.4}&\textbf{57.4}&{99.7}&\textbf{96.7}&\textbf{72.1}&\textbf{88.2}&\textbf{87.8}&\textbf{95.6}&\textbf{87.8}&\textbf{79.2}&\textbf{82.0}&\textbf{80.8}&\textbf{85.7}&\textbf{86.9}&\textbf{97.0}&\textbf{90.4}&\textbf{74.3}&\textbf{86.3}\\
\bottomrule
\end{tabular}
\end{adjustbox}
\label{tab:att}
\vspace{-5pt}
\end{table*}

\textbf{Data Statistics.}  
Fig.~\ref{fig:data_stat} provides key statistics of the \dataset~dataset.
In (a), the data taxonomy organizes the dataset into three primary face perception tasks: Emotion Intelligence, Expression Analysis, and Attribution Perception, covering a wide range of facial features and emotional attributes.
(b) The data distribution plots show the relative face area and video duration across different datasets, illustrating the diversity and variety of video data present in \dataset.
(c) The annotation distribution includes the breakdown of facial views (head, half, full) and video length, accompanied by a word cloud highlighting the most frequently annotated terms, such as ``neutral'', ``face'', and ``expression''.
(d) Data statistics compares \dataset~with other emotion and video datasets, showing that \dataset~provides a richer set of annotations and label types, including fine-grained emotion, rationales, and comprehensive video data, making it a unique and valuable resource for emotion-centric research. More details can be refer to Sec.~\ref{sec:Dataset Details} of appendix.

\section{Experiment}
\label{sec:exp}

As shown in Tab.~\ref{tab:att} and ~\ref{tab:emo}, we conduct  experiments to verify the effectiveness of \model~on three types of tasks: attribute perception, expression analysis, and emotion understanding.
We compare \model~with 5 closed MLLM APIs and 13 open-sourced VideoLLMs with scales ranging from 1B to 8B.
More training details (Sec.~\ref{sec:imple}) and evaluation settings (Sec.~\ref{sec:benchmark}) please refer to appendix.

\subsection{SOTA Comparison}
We benchmark \model~on 40 metrics, spanning 6 closed-set attribute tasks, 12 open-set attribute tasks, 9 expression tasks, and 6 high-level emotion understanding tasks.

\textbf{Scale}:
Our models significantly outperform existing closed and open-source VideoLLMs across all metrics from 1B to 8B scales.
At the 1-3B / 7-8B scale, our VidEmo-Base model (3B/7B) achieves an overall average accuracy of 62.4\%/64.1\%, outperforming the strongest baseline, Qwen2.5-VL at 46.1\%/51.7\%, by a margin of +16.3\%/+12.4\%.
The consistent improvement across scales demonstrate the effectiveness of our proposed curriculum learning as well as affective-tree reward in scaling up our foundation model.

\textbf{Attribute \& Expression \& Emotion Tasks}:
We conduct a comprehensive analysis of \model~across three core task categories in the Emo-CFG benchmark: attribute perception, expression analysis, and emotion understanding.
In attribute tasks, which include both closed-set (e.g., identity, head pose) and open-set (e.g., hair type, age, skin tone) recognition, VidEmo\ achieves an average score of 86.3\%, surpassing all baselines including Qwen2.5-VL (7B) at 80.6\%, yielding a +5.7\% improvement. Particularly, our model achieves 99.7\% on identity verification, 95.6\% on facial shape, and 97.0\% on gender prediction, reflecting strong generalization on fine-grained visual perception.
In expression analysis, covering single-label, multi-label, and fine-grained classification, as well as micro-expression and AU detection, VidEmo delivers an average of 39.9\%, outperforming Qwen2.5-VL (7B) by +6.8\%. Notably, VidEmo leads in fine-grained expression classification (35.6\% vs. 29.7\%) and micro-expression detection (20.4\% vs. 13.6\%), demonstrating its sensitivity to subtle and transient affective cues.
In emotion understanding tasks—spanning instruction-following, fluency, response accuracy, and video-text relevance—VidEmo\ achieves 69.3\% on average, outperforming all prior models, including Gemini 2.0 (63.5\%) and Qwen2.5-VL (7B) (63.6\%), with improvements of over +5\%. It sets new benchmarks on tasks like instruction adherence (68.1\%), fluency (69.1\%), and video-text relevance (69.3\%), showcasing its capacity for coherent, explainable, and semantically grounded emotional inference.
We also notice that for the attribute tasks achieves an higher averaged performance.
This finding also aligns with the dynamics we observed during training, that the perplexity of model increase with a sequential order of attribute, expression, and emotion.
\begin{table}[!tbp]
\centering
\tabcolsep=2.5pt
\caption{\textbf{Comparison with 18 leading VideoLLMs on 11 expression analysis tasks and 6 fine-grained emotion understanding tasks of~\dataset}. Sin: single-label classification, Mul: multi-label classification, Fine: fine-grained classification, Mic: micro-expression detection, AU: action unit detection, Cap: caption,  Conv: conversation emotion analysis, VTR: video-text relevance, Flu: fluency, RA: response accuracy, IA: instruction adherence, Clu: clue overlap, Lab: label overlap, AVG: average.
}
\vspace{-5.5pt}
\begin{adjustbox}{width=\textwidth}
\begin{tabular}{p{3.05cm}p{.4cm}<{\centering}p{.61cm}<{\centering}p{.61cm}<{\centering}p{.61cm}<{\centering}p{.61cm}<{\centering}p{.61cm}<{\centering}p{.61cm}<{\centering}p{.61cm}<{\centering}p{.61cm}<{\centering}p{.61cm}<{\centering}p{.61cm}<{\centering}p{.61cm}<{\centering}p{.61cm}<{\centering}p{.68cm}<{\centering}|p{.61cm}<{\centering}p{.61cm}<{\centering}p{.61cm}<{\centering}p{.61cm}<{\centering}p{.61cm}<{\centering}p{.61cm}<{\centering}p{.61cm}<{\centering}}
\toprule
\multirow{2}{*}{Model} &\multirow{2}{*}{Size} & \multicolumn{3}{c}{Emotion} & \multirow{2}{*}{AVG} & \multicolumn{2}{c}{Sentiment} & \multirow{2}{*}{AVG} & \multicolumn{2}{c}{Cues} & \multirow{2}{*}{AVG} & \multicolumn{2}{c}{Complex}  & \multirow{2}{*}{AVG}& \multicolumn{6}{c}{Emotion Understanding}  & \multirow{2}{*}{AVG}\\
\cmidrule(lr){3-5}
\cmidrule(lr){7-8}
\cmidrule(lr){10-11}
\cmidrule(lr){13-14}
\cmidrule(lr){16-21}
& & Sin & Mul & Fine & & Sin & Fine & & Mic & AU &  & Cap & Conv &  & Lab & Clu &IA & RA & VTR & Flu &  \\
\midrule
\rowcolor{gray!10}
\multicolumn{4}{l}{\textit{Closed MLLM API}}
            &    &    &    &    &    &    &    &    &    &    &    &    &    &    &    &    &    &    \\
Gemini 2.0~\cite{team2024gemini}
&-&45.3&42.2&23.6&37.0&45.0&26.0&35.5&16.2&36.1&26.1&42.0&50.6&46.3&51.0&52.6&58.2&60.2&66.8&92.1&63.5\\
Claude 3 Sonnet~\cite{anthropic2024claude}
&-&26.6&17.6&16.6&20.3&37.6&23.3&30.4&29.2&37.1&33.2&50.3&36.3&43.3&47.5&49.5&57.6&54.9&62.0&92.5&60.7\\
Qwen-VL-MAX~\cite{qwen2023qwen}
&-&29.3&25.8&24.6&26.6&32.6&21.3&27.0&11.7&24.0&17.9&46.4&49.3&47.8   &47.9&50.7&57.4&56.6&63.6&92.0&61.4\\
GPT-4o~\cite{OpenAI2024}
&-&20.6&19.0&04.3&14.6&29.6&18.0&23.8&29.6&02.3    &16.0    &38.1&43.6&40.8&30.0&35.4&40.8&43.2&50.9&87.4&48.0\\
GPT-4o mini~\cite{OpenAI2024}               
&-
&19.6&20.2&07.3&15.7&29.9&16.6&23.3&21.9&08.9&15.4&40.7&47.6&44.2 &35.5&39.0&46.2&46.7&54.4&91.0&52.1 \\

\midrule
\rowcolor{gray!10}
\multicolumn{4}{l}{\textit{Open-sourced 1-3B Video MLLM}}
            &    &    &    &    &    &    &    &    &    &    &    &    &    &    &    &    &    &    \\
LLaVA-OV~\cite{li2024llava}
&1B&28.3&17.5&05.3&17.0&36.3&15.0&25.6&13.4&15.8&14.6&40.2&29.6&34.9&17.2&18.2&28.6&23.5&28.8&89.0&34.2\\
InternVL2.5~\cite{chen2024expanding}
&2B&23.0&16.1&08.3&15.8&31.6&15.6&23.6&09.7&07.4&08.5&43.6&26.0&34.8&40.1&42.9&51.6&49.6&56.4&90.8&55.2\\
VideoLLaMA3~\cite{zhang2025videollama}
&2B&29.3&18.3&11.6&19.7&34.3&16.0&25.1&06.5&06.5&06.5&20.1&31.3&25.7&24.0&22.4&29.5&33.4&40.3&86.8&39.4\\
mPLUG-Owl3~\cite{ye2024mplug}
&2B&32.3&18.4&05.0&18.6&35.6&07.6&21.6&\textbf{28.0}&22.7&25.3&45.0&34.3&39.7&27.8&27.9&35.8&34.6&41.3&89.9&42.9\\
Qwen2.5-VL~\cite{qwen2025qwen}
&3B&36.0&23.7&09.3&23.0&36.6&20.3&28.5&12.1&23.0&17.6&44.2&40.3&42.2&38.2&40.9&49.7&49.2&56.3&91.4&54.3\\
\midrule
\rowcolor{LightCyan}
VidEmo-Base
&3B&\textbf{46.0}&\textbf{38.0}&\textbf{26.0}&\textbf{36.6}    &\textbf{40.3}&\textbf{32.6}&\textbf{36.5}&22.3&\textbf{32.1}&\textbf{27.2}&\textbf{48.1}&\textbf{42.0}&\textbf{45.0} &\textbf{57.3}&\textbf{59.6}&\textbf{70.7}&\textbf{62.7}&\textbf{68.1}&\textbf{93.1}&\textbf{68.6}\\ 

\midrule
\rowcolor{gray!10}
\multicolumn{4}{l}{\textit{Open-sourced 7B+ Video MLLM}}
            &    &    &    &    &    &    &    &    &    &    &    &    &    &    &    &    &    &    \\
ShareGPT4Video~\cite{chen2024sharegpt4video}
&8B&07.6&06.0&04.6&06.1&38.0&14.3&26.1&09.7&01.4&05.6&46.2&32.3&39.2&16.1&18.8&34.4&21.8&26.0&91.1&34.7\\
InternVL2.5~\cite{chen2024expanding}
&8B&28.0&26.2&09.0&21.0&29.3&18.3&23.8&16.2&12.8&14.5&40.8&35.0&37.9&52.3&53.4&61.5&59.8&66.1&92.4&64.2\\
LLaVA-N-Video~\cite{liu2024llavanext}
&7B&24.3&23.7&10.6&19.5&39.0&14.0&26.5&10.9&13.3&12.1&44.1&39.0&41.5&33.7&33.2&43.3&38.8&45.2&90.9&47.5\\
LLaVA-OV~\cite{li2024llava}
&7B&31.6&22.7&10.3&21.5&36.0&20.0&28.0&11.7&15.5&13.6&42.2&43.0&42.6&36.5&39.3&49.5&46.2&53.1&91.0&52.6\\
VideoLLaMA3~\cite{zhang2025videollama}
&7B&27.6&23.9&10.6&20.7&31.0&19.3&25.1&08.9&10.8&09.8&33.2&36.0&34.6&42.2&44.4&53.1&52.9&59.5&89.5&56.9\\
LLaVA-Video~\cite{zhang2024video}
&7B&32.6&22.9&09.3&21.6&35.3&20.6&28.0&12.1&02.5&07.3&45.8&42.0&43.9&38.8&40.9&50.1&47.8&54.6&91.2&53.9\\
mPLUG-Owl3~\cite{ye2024mplug}
&7B&30.0&22.2&10.8&21.0&29.3&15.6&22.5&\textbf{21.5}&25.1&23.3&47.2&33.6&40.4&36.5&35.6&44.2&43.8&50.3&90.8&50.2\\
Qwen2.5-VL~\cite{qwen2025qwen}
&7B&38.6&27.0&12.3&26.0&30.0&22.3&26.1&10.5&14.4&12.5&46.2&44.3&45.2&50.7&52.1&60.0&59.7&66.3&92.7&63.6\\
\midrule
\rowcolor{LightCyan}
\model-Base
&7B&47.3&37.6&34.6&39.8&\textbf{42.3}&36.0&39.1&18.2&\textbf{34.2}&26.2&50.0&48.6&49.3&55.9&57.4&67.9&62.6&68.3&\textbf{92.8}&67.5\\
\rowcolor{LightCyan}
\model-T1
&7B&\textbf{49.7}&\textbf{38.8}&\textbf{35.6}&\textbf{41.3}&\textbf{42.3}&\textbf{37.5}&\textbf{39.9}&20.4&34.1&\textbf{27.3}&\textbf{50.7}&\textbf{52.9}&\textbf{51.8}&\textbf{59.3}&\textbf{61.2}&\textbf{68.1}&\textbf{65.9}&\textbf{69.1}&{92.6}&\textbf{69.3}\\
\bottomrule
\end{tabular}
\end{adjustbox}
\vspace{-5pt}

\label{tab:emo}
\end{table}

\subsection{Discussion}

\textbf{Open-sourced Models and Closed Models}:
We evaluate both open-sourced and closed multimodal large models (MLLMs) on the Emo-CFG benchmark. Closed models, including Gemini 2.0, Claude 3 Sonnet, GPT-4o, GPT-4o mini, and Qwen-VL-MAX, typically operate as APIs with proprietary architectures and in-distribution training data. In contrast, open-sourced models span both small and large-scale variants (1–7B parameters), including LLaVA-OV, InternVL2.5, VideoLLaMA3, mPLUG-Owl, Qwen-VL, and our own VidEmo series.
Across all three evaluation categories—attribute perception, expression analysis, and emotion understanding—our open-sourced VidEmo-T1 (7B) outperforms all closed-source models. For instance, VidEmo-T1 achieves 86.3\% in attribute perception, surpassing Gemini 2.0 by +9.8\%, and obtains 39.9\% on expression tasks, outperforming Claude 3 by +16.6\%. Notably, in high-level emotion understanding, VidEmo-T1 reaches 69.3\%, exceeding GPT-4o’s 48.0\% by a margin of +21.3\%.

\textbf{Base Model and Reasoning Model}
We further compare our base model (VidEmo-Base, 7B) with the reasoning-enhanced model (VidEmo-T1) to assess the effectiveness of affective-tree reasoning.
In attribute perception, VidEmo-T1 improves the average performance from 84.7\% to 86.3\%, with notable gains in key tasks such as head pose estimation (+1.9\%, from 93.4 to 96.7), facial feature prediction (+1.3\%, from 85.6 to 86.9), and gender recognition (+1.9\%, from 89.5 to 90.4).
In expression analysis, VidEmo-T1 shows consistent improvements across all sub-tasks. The average increases from 39.1\% to 41.3\%, with gains in micro-expression detection (+2.2\%) and fine-grained expression recognition (+1.0\%). 
Most notably, in emotion understanding, VidEmo-T1 achieves a substantial improvement from 67.5\% to 69.3\%, with strong gains in fluency (+2.3\%, from 67.9 to 70.2), video-text relevance (+2.6\%), and instruction adherence (+1.9\%). 
\begin{wraptable}{r}{0.45\textwidth}
\tabcolsep=3pt
\vspace{-3pt}
\caption{\textbf{Downstream tasks on DFEW and MAFW datasets.}}
\vspace{2pt}
\begin{adjustbox}{width=\linewidth}
\begin{tabular}
{p{3cm}p{.8cm}<{\centering}p{.8cm}<{\centering}p{.8cm}<{\centering}p{.8cm}<{\centering}}
\toprule
\multirow{2}{*}{Model} & \multicolumn{2}{c}{DFEW}  & \multicolumn{2}{c}{MAFW}\\
\cmidrule(lr){2-3}\cmidrule(lr){4-5}
&UAR&WAR&UAR&WAR\\
\midrule
EmotionCLIP~\cite{zhang2023learning} & 13.77 & 19.89 & 9.20 & 11.65 \\
Exp-CLIP~\cite{zhao2025enhancing} & 24.25 & 25.87 & 17.53 & 20.27 \\
EmoCLIP~\cite{foteinopoulou2024emoclip} & 36.76 & 46.27 & 25.86 & 33.49 \\
EmoCapCLIP~\cite{sun2025learning} & 42.19 & 43.99 & 30.85 & 34.50 \\
I3D~\cite{carreira2017quo} & 46.52 & 58.27 & - & - \\
F-DFER~\cite{zhao2021former} & 53.69 & 65.70 & - & - \\
EST~\cite{liu2023expression} & 53.43 & 65.85 & - & - \\
IAL~\cite{li2023intensity} & 55.71 & 69.24 & - & - \\
CLIPER~\cite{li2024cliper} & 57.56 & 70.84 & - & - \\
DFER-CLIP~\cite{zhao2023prompting} & 59.61 & 71.25 & 38.89 & 52.55 \\
EMO-LLaMA~\cite{xing2024emo} & 60.23 & 65.89 & 41.57 & 48.63 \\
\midrule
\rowcolor{LightCyan}
VidEmo (Ours) & \textbf{64.92} & \textbf{73.10} & \textbf{44.02} & \textbf{54.86} \\
\bottomrule
\end{tabular}
\end{adjustbox}
\vspace{-15pt}
\label{tab:specific}
\end{wraptable}

\textbf{Downstream Tasks.}
For downstream emotion-related tasks, we evaluate the capacity of VidEmo on facial expression recognition using the DFEW and MAFW datasets.
When fine-tuned for specific tasks, VidEmo consistently outperforms both traditional video-based expression recognition methods and zero-shot emotion-oriented CLIP approaches.
As shown in Table~\ref{tab:specific}, VidEmo achieves a performance gains compared to state-of-the-art baselines (EMO-LLaMA), with an average improvement of 9.4\% over the previous best results across both datasets.
Notably,VidEmo achieves relative improvements of 7.8\% in UAR and 10.9\% in WAR on DFEW, and 5.9\% in UAR and 12.8\% in WAR on MAFW, demonstrating its superior effectiveness for facial expression recognition tasks.
\subsection{Ablation Study}

To investigate the effectiveness of curriculum emotion learning (CEL), affective-tree reward (ATR), and emotion reasoning (ER), we conduct a component-wise ablation study, as presented in Tab.~\ref{tab:ablation} and more in-depth analysis in Tab.~\ref{tab:cel}, Tab.~\ref{tab:atr} and Tab.~\ref{tab:er} of appendix.
In this study, we assess the contribution of each component by analyzing the performance of the model under different configurations, removing one or more of the components.
With the ablation studies conducted, we find four interesting observations:

\begin{itemize}[leftmargin=15pt]
\item When none of the components are used, the model achieves an average performance of 51.4.
This baseline highlights the importance of incorporating these components into the model for improved performance.

\item With the inclusion of CEL alone, the model performance increases to an average of 61.9, demonstrating the positive impact of curriculum emotion learning on the model’s ability to handle emotional contexts.
Specifically, we observe improvements in the emotion-related metrics, particularly in the expression and emotion attributes.

\item Introducing ATR alongside CEL further enhances the model's performance, with an average score of 63.6.
The inclusion of ATR results in more refined emotion handling, as seen in the improvements in the emotion and expression attributes.

\item The full model, with CEL, ATR, and ER, achieves the highest performance, with an average score of 67.0.
This configuration benefits from the combined effects of all components, especially in emotion reasoning, where the model shows notable improvements across all attributes, particularly in the expression and emotion metrics.
\end{itemize}

\subsection{Dataset Verification}
As data scale increases beyond 50K samples, maintaining consistent data quality becomes challenging.
Our data pipeline offers a systematic solution to this problem.
To assess the quality of the generated expressions, we conducted a user study on a manually inspected subset of test samples to verify their alignment with the intended emotional semantics.
Specifically, we compared Emo-CFG with CelebV-Text, the largest human-labeled video emotion dataset, across three key dimensions: precision, rationality, and complementarity.
Preference rates for Emo-CFG across these dimensions reached 95\%, 92\%, and 93\%, respectively, with statistically significant differences (Wilcoxon signed-rank test, $p<0.01$).
This result demonstrates that Emo-CFG provides more precise and expressive emotional representations than existing benchmarks.
\begin{table}[!tbp]
\begin{minipage}[t]{0.37\textwidth}
\makeatletter\def\@captype{table}
\centering
\tabcolsep=3pt
\caption{\textbf{Ablation study on the proposed components for our \model.} CEL = curriculum emotion learning, ATR = affective-tree reward, ER = emotion reasoning.}
\begin{adjustbox}{width=\textwidth}
\begin{tabular}
{p{.6cm}<{\centering}p{.6cm}<{\centering}p{.6cm}<{\centering}p{.6cm}<{\centering}p{.6cm}<{\centering}p{.6cm}<{\centering}p{.6cm}<{\centering}p{.6cm}<{\centering}}
\toprule
\multicolumn{3}{c}{Components}&\multirow{2}{*}{Att}&\multirow{2}{*}{Exp}&\multirow{2}{*}{Emo}&\multirow{2}{*}{Avg}\\
\cmidrule(lr){1-3}
CEL&ATR&ER\ \\
\midrule
&&&
63.5&27.3&63.6&51.4\\
\checkmark&&&
79.5&38.7&67.5&61.9\\
\rowcolor{LightCyan}
\checkmark&\checkmark&&
81.3&40.1&69.3&63.6\\
\checkmark&\checkmark&\checkmark&
84.5&43.8&72.9&67.0\\
\bottomrule
\end{tabular}
\end{adjustbox}
\label{tab:ablation}
\end{minipage}
\hspace{.005\textwidth}
\begin{minipage}[t]{0.62\textwidth}
\makeatletter\def\@captype{table}
\centering
\tabcolsep=3pt
\caption{\textbf{User study across three dimensions between \dataset~and CelebV-Text.} We evaluate the label quality on precision, rationality, and complementary through pairwise comparison with 50 videos and 25 users. All three dimensions show statistically significant preference for \dataset.}
\begin{adjustbox}{width=\textwidth}
\begin{tabular}
{p{2.2cm}<{\centering}p{.6cm}<{\centering}p{.6cm}<{\centering}p{.6cm}<{\centering}p{.6cm}<{\centering}p{.6cm}<{\centering}p{1.2cm}<{\centering}p{1.2cm}<{\centering}p{1.8cm}<{\centering}}
\toprule
\multirow{2}{*}{Dimension} & \multicolumn{3}{c}{Pairwise} & \multirow{2}{*}{\#Vid} & \multirow{2}{*}{\#Usr} & \multirow{2}{*}{Prefer} & \multirow{2}{*}{p-value} \\
\cmidrule(lr){2-4}
& Win & Tie & Loss\\
\midrule
Precision & 964 & 204 & 82 & 50 & 25 & 95.5\% & 0.00021 \\
Rationality & 1082 & 87 & 81 & 50 & 25 & 92.1\% & 0.00015 \\
Complementary & 1172 & 23 & 55 & 50 & 25 & 93.0\% & 0.00008 \\
\bottomrule
\end{tabular}
\end{adjustbox}
\label{tab:user_study}
\end{minipage}
\end{table}

\section{Conclusion}
\label{sec:conclusion}
In this work, we introduced \model, a family of video-based emotion foundation models designed to unify fine-grained facial attribute perception, expression analysis, and high-level emotion understanding. Our framework integrates curriculum emotion learning with a novel affective-tree reasoning paradigm, enabling interpretable and structured emotion inference. We further curated \dataset, a large-scale, instruction-driven dataset with hierarchical annotations and rationale-grounded data, which serves as a foundamental data infrastructure for training and evaluation. Experimental results on the Emo-CFG benchmark demonstrate that \model~consistently outperforms existing open- and closed-source VideoLLMs across 15 tasks, setting up a new milestone in all the attribute perception, expression analysis, and emotion understanding tasks.

\textbf{Limitations.}
While \model~exhibits strong generalization across diverse tasks, several limitations remain.
First, like most existing VideoLLMs, \model~is susceptible to generating counterfactual content, which can lead to false narratives or emotionally inconsistent descriptions.
Second, emotion understanding is inherently multimodal; integrating additional modalities such as audio or contextual cues could significantly enrich affective reasoning and we view \model~as a strong foundation for future work in this direction, enabling the exploration of richer, more holistic emotion understanding.

\textbf{Acknowledgements.}
This work was supported by the National Natural Science Foundation of China (No.623B2056), the Natural Science Foundation of Tianjin, China (No.24JCZXJC00040), the Fundamental Research Funds for the Central Universities, the Supercomputing Center of Nankai University (NKSC).
We sincerely thank the reviewer team (KQSY, 96WN, ntby, and JqQW) for their invaluable feedback to improve our manuscript.
{
\small
\bibliographystyle{plain}
\bibliography{main}
}

\clearpage

\appendix

\section{Implementation and Training Details}
\label{sec:imple}
Following the approach outlined by Qwen2.5-VL~\cite{qwen2025qwen}, we adopt a Vision Transformer (ViT)-based architecture for the visual encoder and utilize an autoregressive model for the text encoder.
For the foundational large language model (LLM), we select models within the 3B to 7B parameter range.
\model~is pre-trained for 3 epochs with a batch size of 1024 and is subsequently post-trained for 1 epoch with a batch size of 128.
We employ the AdamW~\cite{loshchilov2018decoupled} optimizer with a cosine learning rate schedule.
The learning rate is set to 2e-5 for SFT stage and 1e-5 for RL stage, with a warmup rate of 0.03.
As shown in Table~\ref{tab:details}, we follow the existing MLLM training setting and use a two-stage tuning paradigm.
\begin{itemize}[leftmargin=15pt]
\item \textbf{Backbone}: We use the same backbone LLM and vision encoder as Qwen2.5-VL.
\item \textbf{Data}: We train VidEmo by our constructed Emo-CFG and textual knowledge dataset MAGPIE.
\item \textbf{Hyperparameter}: We follow the common setting and train the model by default learning rate, weight decay, and batch size.
\end{itemize}
\begin{table}[hbp]
\centering
\caption{\textbf{Implementation details and hyperparameters for our~\model~family.}}
\begin{adjustbox}{max width=\textwidth}
\begin{tabular}{lcccccccc}
\toprule
\multirow{2}{*}{\textbf{Model}} & \multicolumn{2}{c}{\textbf{Backbone}} & \multicolumn{2}{c}{\textbf{Data}} & \multicolumn{4}{c}{\textbf{Hyperparameter}} \\
\cmidrule(lr){2-3}
\cmidrule(lr){4-5}
\cmidrule(lr){6-9}
 & LLM & Vision & SFT & RL & lr & wd & bs & rollout  \\
\midrule
\model-Base-3B & Qwen2.5-3B & ViT & \dataset& - & 2e-5 & 0 & 1024 & -\\
\model-Base-7B & Qwen2.5-7B & ViT & \dataset& - & 2e-5 & 0 & 1024 & - \\
\model-T1-7B & Qwen2.5-7B & ViT & \dataset& \dataset & 1e-5 & 0 & 128 & 8 \\

\bottomrule
\end{tabular}
\end{adjustbox}
\label{tab:details}
\end{table}

\section{\dataset~Dataset Details}
\label{sec:Dataset Details}
\subsection{Dataset Construction}
In this section, we provide an overview of the training data used in our \dataset\ dataset, which is sourced from multiple datasets to address various tasks related to emotion and attribute perception.
The data is collected to support model training, and the details of each task are summarized in Table~\ref{tab:emotion_tasks_train}.
We illustrate the scenario, data source, task, data number, and ratio of our training data.
\begin{table*}[ht]
\small
\centering
\caption{The overview of our training data. All the data used for training are sampled from the training or validation split of the source datasets. QA: Question Answering. OPEN: Open-ended Question Answering. SR: Sentiment Recognition. ER: Emotion Recognition.} 
\resizebox{\linewidth}{!}{
\begin{tabular}{lcccp{3cm}<{\centering}}
\toprule
\textbf{Scenario} & \textbf{Data Source} & \textbf{Task} & \textbf{Number} & \textbf{Ratio} \\
\hline\midrule
\rowcolor{gray!25}\multicolumn{5}{c}{\textbf{Attribute Perception}} \\\midrule
Appearance Recognition  & CelebV-HQ     & Multi-label QA    &32010  &8.16\% \\
Appearance Caption      & CelebV-Text   & Caption              &59879  &15.26\% \\
Action Recognition      & CelebV-HQ     & Multi-label QA    &32010  &8.16\% \\
Action Caption          & CelebV-Text   & Caption              &59797  &15.24\% \\
Human Identity          & MEAD          & QA                &19998  &5.10\% \\
Head Pose               & MEAD          & QA                &188590 &48.10\%  \\
\hline\midrule
\rowcolor{gray!25}\multicolumn{5}{c}{\textbf{Open Attribute Perception}} \\
\midrule
Eye             & \multirow{14}{*}{\dataset} & QA \& OPEN & 72197&5.94\% \\
Eyebrow         &                           & QA \& OPEN & 71793&5.90\% \\
Mouth           &                           & QA \& OPEN & 94722&7.79\%\\
Nose            &                           & QA \& OPEN & 126248&10.38\% \\
Hair            &                           & QA \& OPEN & 154994&12.75\% \\
Chin            &                           & QA \& OPEN & 20559&1.69\% \\
Shape           &                           & QA \& OPEN & 78689&6.47\% \\
Feature         &                           & QA \& OPEN & 92830&7.63\% \\
Accessory       &                           & QA \& OPEN & 56835&4.67\%\\
Age             &                           & QA \& OPEN &145359&11.96\% \\
Gender          &                           & QA \& OPEN & 151901&12.49\% \\
Skin          &                           & QA \& OPEN & 104005&8.55\% \\
Body Action     &                           & QA \& OPEN &16577&1.36\%\\
Head Action     &                           & QA \& OPEN &2658&0.22\% \\
Face Action     &                           & QA \& OPEN &26515&2.18\% \\

\hline\midrule
\rowcolor{gray!25}\multicolumn{5}{c}{\textbf{Expression Analysis}} \\\midrule
Single-label SR & MOSEI,MOSI & QA & 19710 &5.66\%  \\
Fine-grained SR & CHSIMSv1  & QA & 1824 &0.52\%  \\
Single-label ER & MAFW,DFEW,MER2023 & QA & 20060 &5.76\%  \\
Multi-label ER & MAFW & Multi-label QA & 7178 &2.06\% \\
Fine-grained ER & MEAD,RAVDESS & QA & 198957 &57.11\%  \\
Micro-expression Detection & CASME,CASME$^2$ & QA & 515 &0.15\%  \\
Action Unit Detection & AffWild2 & QA & 2180 &0.63\%  \\
Conversation Reasoning & PERR,MELD & QA & 38153 &10.95\%  \\
Expression Caption & CelebV-Text  & Caption & 59797 &17.16\%  \\
\hline\midrule
\rowcolor{gray!25}\multicolumn{5}{c}{\textbf{Emotional Intelligent}} \\\midrule
Video-Text Relevance & \multirow{7}{*}{\dataset}  &  & \multirow{6}{*}{78072} &\multirow{6}{*}{39.10\%}  \\
Fluency&&&& \\
Coherency&&Fine-grained&& \\
Response Accuracy&&Caption&& \\
Cue Overlap&&&& \\
Label Overlap&&&& \\
Emotion Reasoning&&Rationale&121618&60.90\% \\
\bottomrule
\end{tabular}}
\label{tab:emotion_tasks_train}
\end{table*}

The attribute perception category includes tasks such as appearance recognition, action recognition, and human identity recognition.
These tasks are sourced from CelebV-HQ, CelebV-Text, and MEAD, with multi-label question answering (QA) and caption generation tasks.
The open attribute perception category involves tasks that focus on the recognition and analysis of open-ended attributes like eye, mouth, nose, shape, gender, and more.
Important attributes such as age, gender, and accessories are covered in this category, and each task plays a significant role in identifying open-ended features that contribute to emotional understanding.

In the expression analysis category, we focus on tasks related to sentiment recognition (SR), emotion recognition (ER), affective cues detection, and complex scenario understanding.
These tasks are sourced from datasets like MOSEI, CHSIMSv1, and CASME and aim to capture fine-grained emotional expressions and actions.
The analysis of micro-expressions and emotion-related cues in this category contributes to the detailed recognition of emotional states.

Lastly, the emotional intelligence category covers important capacities related to video-text relevance, fluency, and emotional reasoning. 
These capacities focus on understanding the relationship between video and text for emotional intelligence applications.
Notably, emotional reasoning tasks make up 60.90\% of the dataset in this category, highlighting the importance of reasoning-based tasks in the overall dataset.

\subsection{Dataset Statistics}

\begin{figure}[ht]
  \centering
  \includegraphics[width=.9\linewidth]{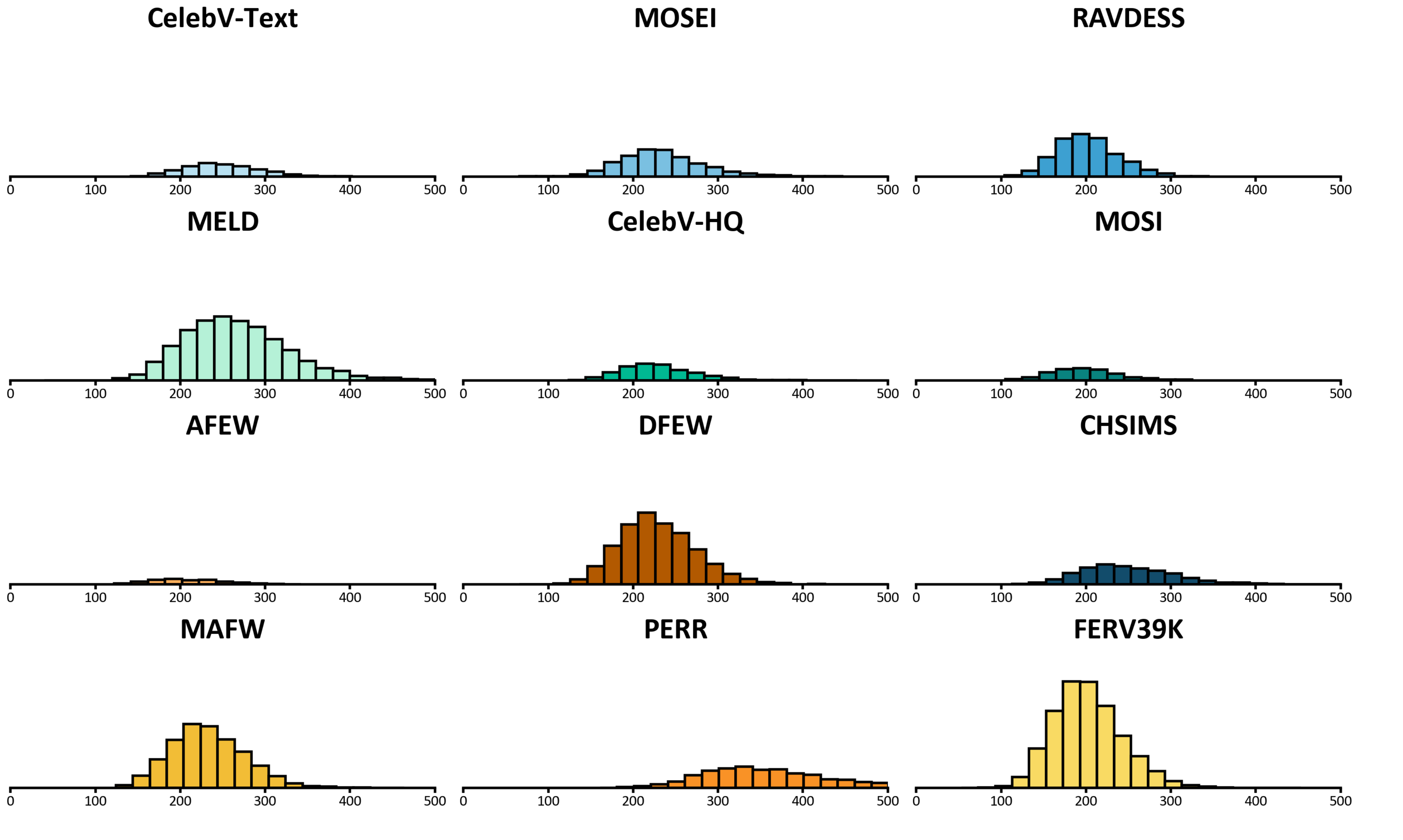}
  \vspace{-5pt}
  \caption{Length Distribution of caption data from the sources of \dataset.}
  \vspace{-5pt}
  \label{fig:data_cap_length}
\end{figure}

\begin{figure}[ht]
  \centering
  \includegraphics[width=.9\linewidth]{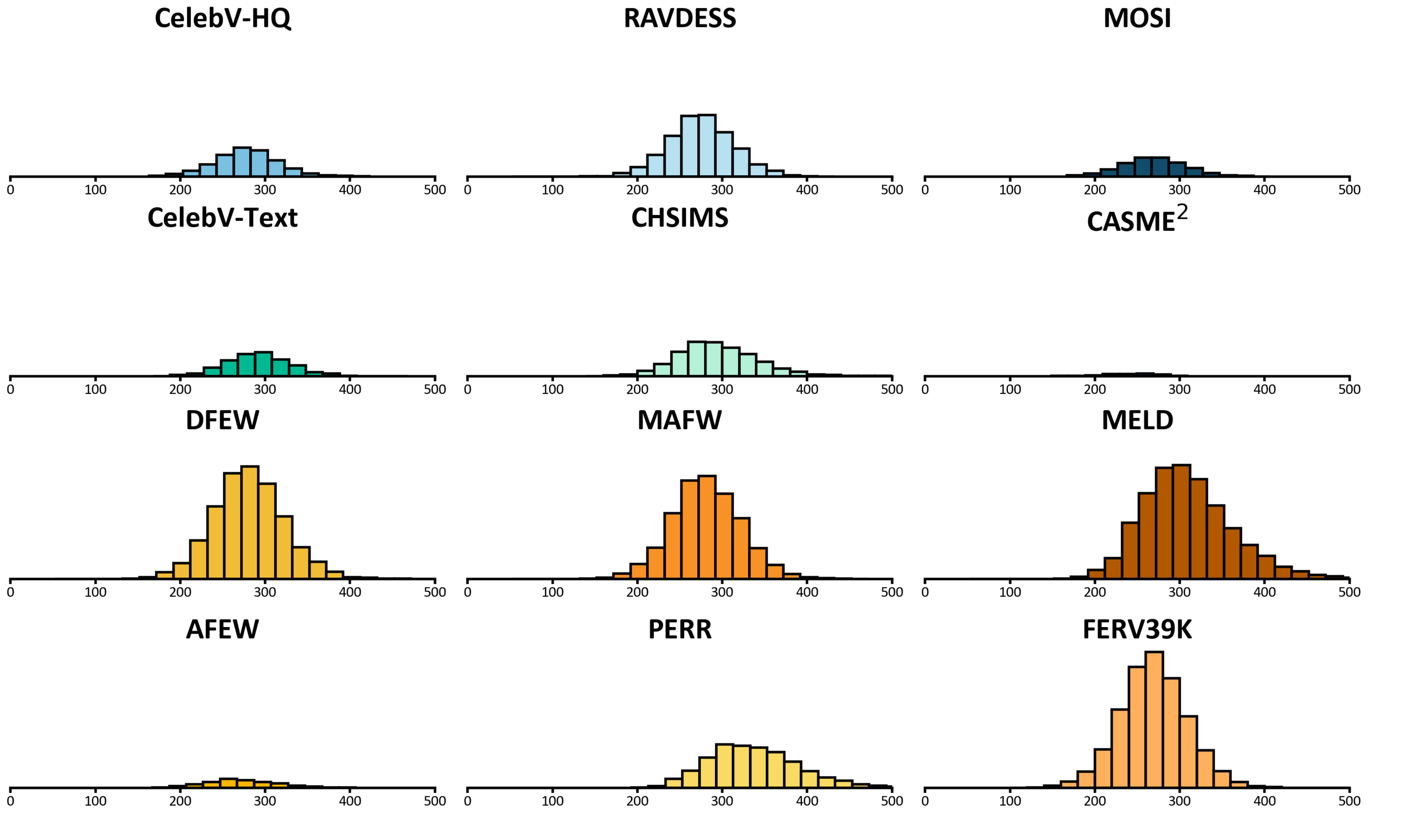}
  \vspace{-5pt}
  \caption{Length Distribution of rationale data from the sources of \dataset.}
  \vspace{-10pt}
  \label{fig:data_rat_length}
\end{figure}

Our constructed \dataset~richs in fine-grained caption for high-level emotion understanding.
We illustrate the caption distribution as shown in Figure~\ref{fig:data_cap_length}.
We displays the distribution of caption lengths across multiple sources, including CelebV-Text, MOSEI, RAVDESS, MELD, CelebV-HQ, MOSI, AFEW, DFEW, CHSIMS, MAFW, PERR, and FERV39K.
Each histogram represents the frequency of captions of varying lengths, ranging from 0 to 500 words.
We can observe that the variability in caption length across these datasets, with some datasets exhibiting a more uniform distribution (e.g., CelebV-Text and MOSI) and others showing skewed distributions (e.g., MAFW and PERR).
We further illustrate the rationale distribution for high-level emotion understanding as shown in Figure~\ref{fig:data_rat_length}.
The rationale lengths exhibit distinct distributions across the datasets.
Some sources, like CelebV-Text and MOSI, show more uniform distributions, while others, such as MAFW and FERV39K, present skewed distributions. 
These distributions are crucial as they reflect the varying complexity and the level of detail involved in the rationales used for emotion analysis and understanding in multimodal tasks.

\section{More Experimental Results}
\label{sec:more_exp}

We further explore the design of the proposed components and analyze their effects in detail.

\textbf{Supervised Finetuning: Curriculum Emotion Learning.}
The effectiveness of our proposed Curriculum Emotion Learning (CEL) approach is demonstrated through an in-depth analysis of the pre-training stages, as outlined in Table~\ref{tab:ablation}.

For the single-stage pre-training, where each stage is isolated, we observe varying levels of performance across different emotion-related tasks.
Specifically, when the model is trained on attribute data alone, the average performance is lower due to the limited complexity of the tasks.
Introducing expression data results in moderate improvements, particularly in expression-related tasks, but the overall performance remains relatively modest (42.2).
The most substantial improvement occurs when emotion data is introduced, leading to a noticeable boost in emotion-related tasks (53.9), but still lacking the holistic integration seen in multi-stage training.

For the our proposed multiple-stage pre-training, where the model is exposed to progressively more complex tasks across all three stages, the performance improves significantly.
The introduction of both attribute and expression data at earlier stages (52.6 average) enables the model to better integrate and align emotion-related information.
With the complete multi-stage pre-training, which includes all three data types, the model achieves a robust performance (61.9 average), indicating that the curriculum learning strategy successfully enhances the model’s understanding of emotional complexity.
The analysis reveals that progressively increasing task difficulty facilitates the model's ability to learn emotion-based tasks more effectively, aligning with our goal of gradually injecting emotion knowledge into the base model.

\begin{table}[ht]
\centering
\tabcolsep=3pt
\caption{\textbf{In-Depth Analysis on curriculum emotion learning with data curation for our \model.} Attr = Attribution Data, Exp = Expression Data, Emo = Emotion Data.}
\begin{tabular}
{p{1.4cm}<{\centering}p{1.4cm}<{\centering}p{1.4cm}<{\centering}ccccc}
\toprule
\multicolumn{3}{c}{CEL}&\multirow{2}{*}{Attribute}&\multirow{2}{*}{Expression}&\multirow{2}{*}{Emotion}&\multirow{2}{*}{Average}\\
\cmidrule(lr){1-3}
Attr&Exp&Emo\ \\
\midrule
\rowcolor{gray!10}
\multicolumn{7}{c}{\ Single Stage}\\
&&&
63.5&27.3&63.6&51.4\\
\checkmark&&&
65.7&24.1&32.4&40.7\\
&\checkmark&&
52.9&31.6&42.1&42.2\\
&&\checkmark&
64.3&30.2&67.3&53.9\\
\midrule
\rowcolor{gray!10}
\multicolumn{7}{c}{\ Multiple Stage}\\
\checkmark&\checkmark&&
77.2&35.1&45.5&52.6\\
\rowcolor{LightCyan}
\checkmark&\checkmark&\checkmark&
79.5&38.7&67.5&61.9\\

\bottomrule
\end{tabular}
\label{tab:cel}
\end{table}

\textbf{Post-training with Affective-Tree Reward.}
The post-training phase with Affective-Tree Reward (ATR) is designed to enhance the model’s performance in emotional reasoning tasks. Table~\ref{tab:atr} presents the results of integrating various components into the post-training process.
Initially, when only the group relative policy optimization (GRPO) is applied, the model achieves an average score of 51.2.
This score serves as the baseline performance before incorporating the Affective-Tree Reward.

Upon adding the Affective-Tree Reward (ATR), the model's performance increases to an average of 61.4, indicating that the inclusion of this reward mechanism improves the model's ability to generate emotion-related captions.
The introduction of the Tree Edit Distance (ATR) further enhances the model’s performance, resulting in an average score of 63.6
This improvement is observed across the attribute, expression, and emotion tasks.
These results demonstrate that the addition of the Affective-Tree Reward and Tree Edit Distance enhances the model's performance in emotional reasoning tasks by improving the accuracy of caption generation and ensuring structural alignment with human-annotated captions.
\begin{table}[ht]
\centering
\tabcolsep=3pt
\caption{\textbf{In-Depth Analysis on affective tree reward for our \model.}}
\begin{tabular}
{p{4.2cm}ccccc}
\toprule
&{Attribute}&{Expression}&{Emotion}&{Average}\\

\midrule
GRPO&
75.7&33.8&44.2&51.2\\
\midrule
\ \ \ + Tree Reward&
80.9&38.1&65.3&61.4\\
\rowcolor{LightCyan}
\ \ \ + Tree Edit Distance (ATR)&
81.3&40.1&69.3&63.6\\

\bottomrule
\end{tabular}
\label{tab:atr}
\end{table}

\textbf{Emotion Reasoning.}
The performance of the model in emotion reasoning tasks is evaluated by varying the number of candidate responses sampled during inference.
As shown in Table~\ref{tab:er}, the baseline model, which uses the Affective-Tree Reward (ATR) with a single output (n=1), achieves an average score of 63.6 across the attribute, expression, and emotion tasks.
This baseline represents the basic reasoning model, which is trained with ATR and generates only one response per query.

When the number of candidate outputs is increased, the model's performance improves.
For instance, when two candidate responses (n=2) are sampled, the model achieves an average score of 65.1, with notable improvements in expression and emotion tasks.
Further increasing the number of candidate outputs to four (n=4) results in a slight performance boost, bringing the average score to 66.2.
This trend continues with the sampling of eight candidate responses (n=8), where the model achieves the highest average score of 67.0, along with consistent improvements across all tasks.

This analysis highlights the benefit of the search-based reasoning strategy in enhancing the model's performance.
By sampling multiple candidate outputs and selecting the best one based on a reward-guided scoring mechanism, the model is able to refine its emotional reasoning process.
\begin{table}[ht]
\centering
\tabcolsep=3pt
\caption{\textbf{In-Depth Analysis on emotion reasoning for our \model.} Baseline is the basic reasoning model trained with ATR and only outputs one response (n=1).}
\begin{tabular}
{p{4.2cm}ccccc}
\toprule
&{Attribute}&{Expression}&{Emotion}&{Average}\\
\midrule
Baseline&
81.3&40.1&69.3&63.6\\
\midrule
\ \ \ +Emotion Reasoning (n=2)&
82.9&42.3&70.1&65.1\\
\ \ \ +Emotion Reasoning (n=4)&
84.2&43.2&71.2&66.2\\
\rowcolor{LightCyan}
\ \ \ +Emotion Reasoning (n=8)&
84.5&43.8&72.9&67.0\\
\bottomrule
\end{tabular}
\label{tab:er}
\end{table}

\clearpage

\begin{table*}[ht]
\small
\centering
\caption{The overview of our evaluation benchmark. All the data used for evaluation are sampled from the testing split of the source datasets. QA: Question Answering. OPEN: Open-ended Question Answering. SR: Sentiment Recognition. ER: Emotion Recognition.}
\resizebox{\linewidth}{!}{
\begin{tabular}{lcccc}
\toprule
\textbf{Scenario} & \textbf{Data Source} & \textbf{Task} & \textbf{Number} & \textbf{Metric} \\
\hline\midrule
\rowcolor{gray!25}\multicolumn{5}{c}{\textbf{Attribute Perception}} \\\midrule
Appearance Recognition & CelebV-HQ & Multi-label QA & 500 &   F1 \\
Appearance Caption & CelebV-Text & OPEN & 500 & GPT score \\
Action Recognition & CelebV-HQ & Multi-label QA & 500 &   F1 \\
Action Caption & CelebV-Text & OPEN & 500 & GPT score \\
Human Identity & MEAD & QA & 500 & ACC \\
Head Pose & MEAD & QA & 500 & ACC \\
\hline\midrule
\rowcolor{gray!25}\multicolumn{5}{c}{\textbf{Open Attribute Perception}} \\
\midrule
Eye             & \multirow{14}{*}{\dataset} & QA \& OPEN & 300 & ACC \& GPT score \\
Eyebrow         &                           & QA \& OPEN & 300 & ACC \& GPT score \\
Mouth           &                           & QA \& OPEN & 300 & ACC \& GPT score \\
Nose            &                           & QA \& OPEN & 300 & ACC \& GPT score \\
Hair            &                           & QA \& OPEN & 300 & ACC \& GPT score \\
Chin            &                           & QA \& OPEN & 300 & ACC \& GPT score \\
Shape           &                           & QA \& OPEN & 300 & ACC \& GPT score \\
Feature         &                           & QA \& OPEN & 300 & ACC \& GPT score \\
Accessory       &                           & QA \& OPEN & 300 & ACC \& GPT score \\
Age             &                           & QA \& OPEN & 300 & ACC \& GPT score \\
Gender          &                           & QA \& OPEN & 300 & ACC \& GPT score \\
Skin            &                           & QA \& OPEN & 300 & ACC \& GPT score \\
Body Action     &                           & QA \& OPEN & 107 & ACC \& GPT score \\
Head Action     &                           & QA \& OPEN & 4 & ACC \& GPT score \\
Face Action     &                           & QA \& OPEN & 189 & ACC \& GPT score \\
\hline\midrule
\rowcolor{gray!25}\multicolumn{5}{c}{\textbf{Expression Analysis}} \\\midrule
Single-label SR & MOSEI,MOSI & QA & 300 & ACC \\
Fine-grained SR & CHSIMSv1  & QA & 300 & ACC \\
Single-label ER & MAFW,DFEW,MER2023 & QA & 300 & ACC \\
Multi-label ER & MAFW & Multi-label QA & 300 &  F1 \\
Fine-grained ER & MEAD,RAVDESS & QA & 300 & ACC \\
Micro-expression Detection & CASME II & QA & 246 & ACC \\
Action Unit Detection & AffWild2 & Multi-label QA & 87 & F1 \\
Conversation Reasoning & PERR,MELD & QA & 300 &ACC  \\
Emotion Caption & CelebV-Text  & OPEN & 300 & GPT score \\
\hline\midrule
\rowcolor{gray!25}\multicolumn{5}{c}{\textbf{Emotional Intelligent}} \\\midrule
Video-Text Relevance & \multirow{6}{*}{\dataset}  &  & \multirow{6}{*}{2600} &  GPT score \\
Fluency&&&&GPT score \\
Coherency&&Fine-grained&&GPT score \\
Response Accuracy&&Caption \& Rationale&&GPT score \\
Cue Overlap&&&&GPT score \\
Label Overlap&&&&GPT score \\\midrule
\bottomrule
\end{tabular}
}
\label{tab:emotion_tasks_test}
\end{table*}

\section{Evaluation Settings}
\label{sec:benchmark}
\subsection{Task \& Source}
We outline the overview of our evaluation benchmark as shown in Table~\ref{tab:emotion_tasks_test}, including scenario, data source, task, data number, ratio and evaluation metrics.

\textbf{Attribute Perception:} 
CelebV-HQ~\cite{zhu2022celebv} for appearance recognition and action recognition.
CelebV-text~\cite{yu2023celebv} for appearance caption and action caption.
MEAD~\cite{wang2020mead} for head pose estimation and human identity verification.

\textbf{Expression Analysis:} 
MOSEI~\cite{zadeh2018multimodal} and MOSI~\cite{zadeh2016mosi} for single-label sentiment recognition.
CH-SIMSv1~\cite{yu2020ch} for fine-grained sentiment recognition.
MAFW~\cite{liu2022mafw}, DFEW~\cite{jiang2020dfew} and MER2023~\cite{lian2023mer} for single-label emotion recognition.
MAFW~\cite{liu2022mafw} for multi-label emotion recognition.
MEAD~\cite{wang2020mead} and REAVDESS~\cite{livingstone2018ryerson} for fine-grained emotion recognition.
CASME~\cite{yan2013casme}, CASME II~\cite{yan2014casme} and CASME$^2$~\cite{qu2017cas} for micro-expression detection.
Aff-Wild~\cite{kollias2019deep} for AU detection.
MELD~\cite{poria2019meld} and PERR~\cite{gao2021pairwise} for Conversation Reasoning.
CelebV-text~\cite{yu2023celebv} for emotion caption.

\textbf{Emotion Understanding:} 
The annotation for open attribute perception and fine-grained caption in Emo-CFG from 17 source datasets: AFEW~\cite{dhall2011acted}, CAER~\cite{lee2019context}, CASME$^2$~\cite{qu2017cas}, CelebV-HQ~\cite{zhu2022celebv}, CelebV-text~\cite{yu2023celebv}, CHSIMSv1~\cite{yu2020ch}, CHSIMSv2~\cite{liu2022make}, DFEW~\cite{jiang2020dfew}, FERV39K~\cite{wang2022ferv39k}, L-SVD~\cite{emotionnet2023}, MAFW~\cite{liu2022mafw}, MELD~\cite{poria2019meld}, MER2023~\cite{lian2023mer}, MOSEI~\cite{zadeh2018multimodal}, MOSI~\cite{zadeh2016mosi}, PERR~\cite{gao2021pairwise}, REAVDESS~\cite{livingstone2018ryerson}, Aff-Wild~\cite{kollias2019deep}, MEAD~\cite{wang2020mead}.

\subsection{Competitive Alternatives}

\begin{itemize}[leftmargin=15pt]
\item \textbf{Gemini 2.0}
Gemini 2.0 is a cutting-edge closed-source multimodal large language model developed by Google DeepMind.
It is designed to handle both textual and visual inputs, excelling in tasks such as video understanding, summarization, and generation.
\item \textbf{Claude-3.5-Sonnet}
Anthropic's Claude-3.5-Sonnet is a closed-source VideoLLM that builds on the Claude series with enhanced capabilities in video comprehension and interaction.
\item \textbf{GPT-4o / 4o mini}
GPT-4o and its lightweight variant, GPT-4o mini, are closed-source VideoLLMs developed by OpenAI.
These models are optimized for visual understanding tasks, offering a balance between computational efficiency and performance.
\item \textbf{Qwen-VL-Max}
Qwen-VL-Max is designed to process complex video content in conjunction with textual inputs, making it a versatile tool for video summarization, captioning, and question-answering tasks.
\item \textbf{Qwen2.5-VL}
Qwen2.5-VL is an advanced open-source vision-language model that excels in multimodal tasks such as object localization, and long-video comprehension.
Its innovative architecture enables efficient visual recognition and interaction with extended temporal video data. 
\item \textbf{InternVL2.5}
InternVL2.5 pushes the boundaries of open-source multimodal models by introducing advanced scaling strategies across model architecture, diverse video-text datasets, and test-time optimization.
\item \textbf{mPLUG-Owl3}
mPLUG-Owl3 is a cutting-edge multi-modal large language model designed to excel in long image-sequence understanding, including tasks involving lengthy videos and interleaved image-text scenarios.
\item \textbf{VideoLLaMA3}
VideoLLaMA3 adopts a vision-centric training paradigm, emphasizing the use of high-quality image-text data to improve multimodal capabilities.
\item \textbf{LLaVA-OV}
By leveraging insights from data, models, and visual representations, LLaVA-OV achieves significant performance improvements in three major computer vision tasks while enabling strong transfer learning across modalities.
\item \textbf{ShareGPT4Video}
ShareGPT4Video introduces a framework for video understanding and generation by leveraging high-quality captions generated through a specific summary prompt. 
\item \textbf{LLaVA-NeXT-Video}
LLaVA-NeXT-Video is an advanced open-sourced large multimodal model designed for comprehensive video understanding, leveraging interleaved data formats to enhance performance across multi-image, video, and 3D tasks.
\item \textbf{LLaVA-Video}
LLaVA-Video is a video understanding model that processes video sequences using a straightforward approach, supporting both fps and uniform frame sampling.
It is modular and scalable, allowing for efficient training and inference with limited resources, and achieves performance comparable to some 7B models on multiple benchmarks.
\end{itemize}
Table~\ref{tab:model_cards} provides model cards for different MLLMs, including reference papers, parameter scale, and links to pre-trained weights.
\begin{table}[h]
\centering
\caption{Model cards for Open-sourced MLLMs.}
\label{tab:model_cards}
\scalebox{0.79}{
\begin{tabular}{l|c|l}
\toprule
\textbf{Model} & \textbf{Scale} & \textbf{Link} \\
\midrule
LLaVA-OV~\cite{li2024llava}
&1B&\textcolor[rgb]{0.93,0.0,0.47}{\url{https://huggingface.co/llava-hf/llava-onevision-qwen2-0.5b-ov-hf}}\\
InternVL2.5~\cite{chen2024expanding}
&2B&\textcolor[rgb]{0.93,0.0,0.47}{\url{https://huggingface.co/OpenGVLab/InternVL2_5-2B}}\\
VideoLLaMA3~\cite{zhang2025videollama}
&2B&\textcolor[rgb]{0.93,0.0,0.47}{\url{https://huggingface.co/DAMO-NLP-SG/VideoLLaMA3-2B}}\\
mPLUG-Owl3~\cite{ye2024mplug}
&2B&\textcolor[rgb]{0.93,0.0,0.47}{\url{https://huggingface.co/mPLUG/mPLUG-Owl3-2B-241014}}\\
Qwen2.5-VL~\cite{qwen2025qwen}
&3B&\textcolor[rgb]{0.93,0.0,0.47}{\url{https://huggingface.co/Qwen/Qwen2.5-VL-3B-Instruct}}\\
\midrule
ShareGPT4Video~\cite{chen2024sharegpt4video}
&8B&\textcolor[rgb]{0.93,0.0,0.47}{\url{https://huggingface.co/Lin-Chen/sharegpt4video-8b}}\\
InternVL2.5~\cite{chen2024expanding}
&8B&\textcolor[rgb]{0.93,0.0,0.47}{\url{https://huggingface.co/OpenGVLab/InternVL2_5-8B}}\\
LLaVA-NeXT-Video~\cite{liu2024llavanext}
&7B&\textcolor[rgb]{0.93,0.0,0.47}{\url{https://huggingface.co/llava-hf/LLaVA-NeXT-Video-7B-hf}}\\
LLaVA-OV~\cite{li2024llava}
&7B&\textcolor[rgb]{0.93,0.0,0.47}{\url{https://huggingface.co/llava-hf/llava-onevision-qwen2-7b-ov-hf}}\\
VideoLLaMA3~\cite{zhang2025videollama}
&7B&\textcolor[rgb]{0.93,0.0,0.47}{\url{https://huggingface.co/DAMO-NLP-SG/VideoLLaMA3-7B}}\\
LLaVA-Video~\cite{zhang2024video}
&7B&\textcolor[rgb]{0.93,0.0,0.47}{\url{https://huggingface.co/lmms-lab/LLaVA-NeXT-Video-7B}}\\
mPLUG-Owl3~\cite{ye2024mplug}
&7B&\textcolor[rgb]{0.93,0.0,0.47}{\url{https://huggingface.co/mPLUG/mPLUG-Owl3-7B-240728}}\\
Qwen2.5-VL~\cite{qwen2025qwen}
&7B&\textcolor[rgb]{0.93,0.0,0.47}{\url{https://huggingface.co/Qwen/Qwen2.5-VL-7B-Instruct}}\\
\bottomrule
\end{tabular}
}
\end{table}

\subsection{Evaluation Metrics}

We evaluate the performance of the model using various metrics (GPT score, Accurary, and F1 score), depending on the task.
For text-oriented tasks such as Appearance Caption, Action Caption, Emotion Caption, Fine-grained Caption, and Open-ended QA, we employ GPT scores, using GPT-4o to score the predictions based on labels and model responses.
For recognition tasks that involve classification, such as Human Identity, Head Pose, choice-based QA, Single-label Sentiment Recognition (SR), Fine-grained SR, Single-label Emotion Recognition (ER), Fine-grained ER, Micro-expression Detection, and Conversation Reasoning, we use accuracy as the primary evaluation metric. Prior to calculating accuracy, GPT-4o is used to convert the model’s responses into standardized labels to ensure consistency in the evaluation.
For multi-label tasks, including Appearance Recognition, Action Recognition, Multi-label ER, and Action Unit Detection, we utilize the F1-score to evaluate the model's performance, capturing both precision and recall in these multi-label settings.

\subsection{Prompts used for Evaluation}
We utilize three distinct prompts during the evaluation phase, each designed for specific aspects of model performance: F1 score, accuracy evaluation, and GPT score computation.

\begin{figure}[ht]
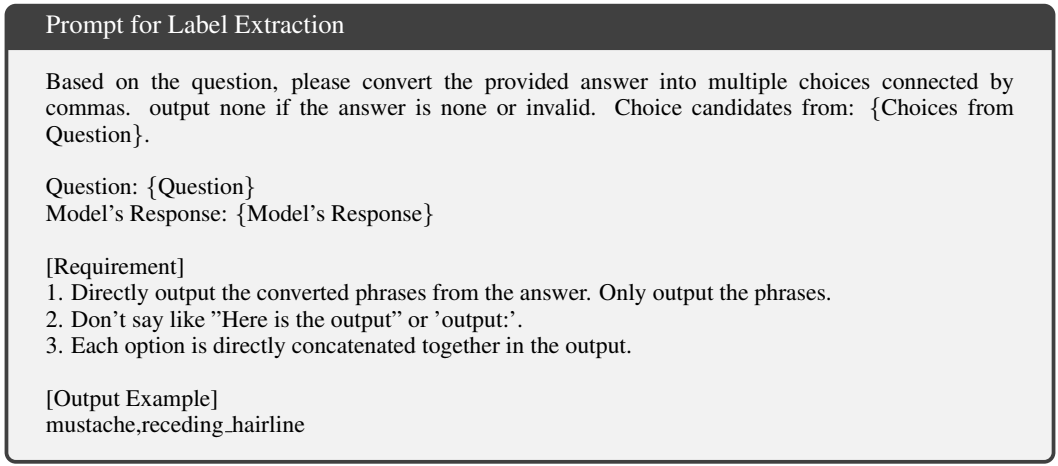
\centering
    \begin{tcolorbox}[title={Prompt for Label Extraction}, fontupper=\footnotesize, before skip=0pt, after skip=0pt]

    Based on the question, please convert the provided answer into multiple choices connected by commas. output none if the answer is none or invalid. Choice candidates from: \{Choices from Question\}. \\

    Question: \{Question\}
    
    Model's Response: \{Model's Response\}\\

    [Requirement]
    
    1. Directly output the converted phrases from the answer. Only output the phrases.
    
    2. Don't say like "Here is the output" or 'output:'. 
    
    3. Each option is directly concatenated together in the output.\\

    [Output Example] 
    
mustache,receding\_hairline

\end{tcolorbox}
\caption{Prompt for extracting multiple phases and compute f1 score from the response of models.}
\vspace{-10pt}
\label{fig:extract label}
\end{figure}

The f1 score prompt, shown in Figure~\ref{fig:extract label}, is used to convert the model's response into multiple choice options. The prompt requires the model to extract relevant labels directly from the provided answer, ensuring that only valid and directly relevant labels are included. If the answer is invalid or none, the prompt instructs the model to output "none." The answer options are then concatenated together, and the model is instructed not to include any additional phrases such as "output:" or "Here is the output."

\begin{figure}[ht]
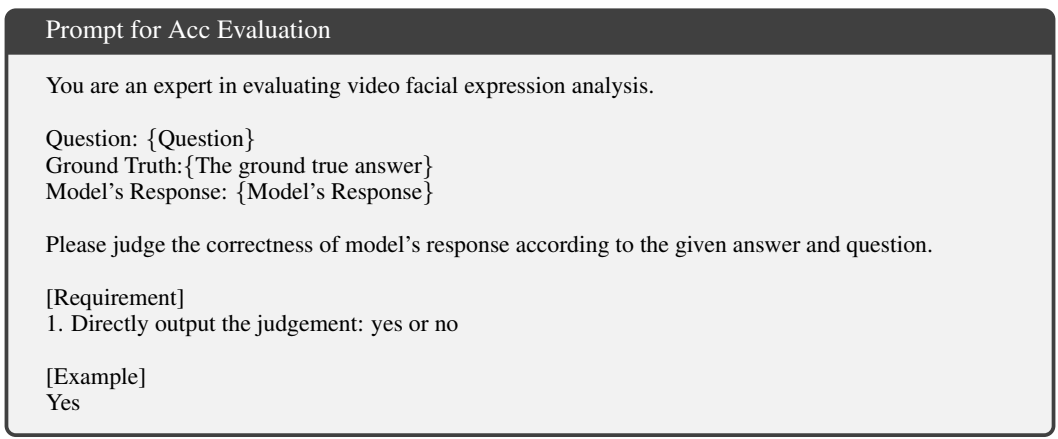
\centering
    \begin{tcolorbox}[title={Prompt for Acc Evaluation}, fontupper=\footnotesize, before skip=0pt, after skip=0pt]

    You are an expert in evaluating video facial expression analysis. \\

    Question: \{Question\}
    
    Ground Truth:\{The ground true answer\}
    
    Model's Response: \{Model's Response\}\\

    Please judge the correctness of model's response according to the given answer and question. \\

    [Requirement]
    
    1. Directly output the judgement: yes or no \\

    [Example] 
    
Yes
\end{tcolorbox}
\caption{Prompt for computing accuracy score.}
\vspace{-10pt}
\label{fig: ACC prompt}
\end{figure}

The accuracy evaluation prompt, depicted in Figure~\ref{fig: ACC prompt}, is designed to assess the correctness of the model’s response. It presents the model with a question, the ground truth answer, and the model's own response. The model is tasked with judging whether the response matches the ground truth and is required to output a simple “yes” or “no” depending on whether the model’s response is correct.

\begin{figure}[ht]
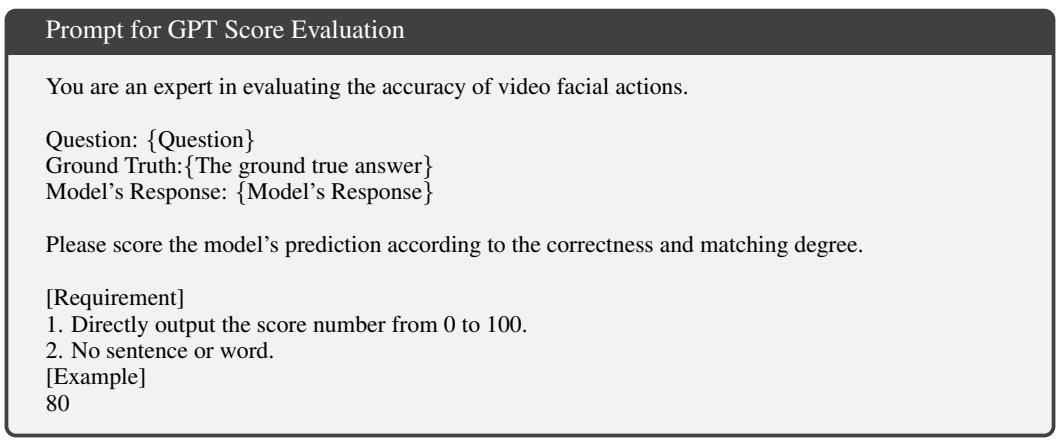
\centering
    \begin{tcolorbox}[title={Prompt for GPT Score Evaluation}, fontupper=\footnotesize, before skip=0pt, after skip=0pt]    
    You are an expert in evaluating the accuracy of video facial actions.\\

    Question: \{Question\}
    
    Ground Truth:\{The ground true answer\}
    
    Model's Response: \{Model's Response\}\\

    Please score the model's prediction according to the correctness and matching degree. \\

    [Requirement] 
    
    1. Directly output the score number from 0 to 100.
    
    2. No sentence or word.

    [Example]
    
80 
\end{tcolorbox}
\caption{Prompt for computing GPT score.}
\label{fig:GPT score}
\vspace{-10pt}
\end{figure}

The GPT score evaluation prompt, illustrated in Figure~\ref{fig:GPT score}, is employed to compute a numerical score reflecting the accuracy and the degree of match between the model’s response and the ground truth. The score is provided as a number between 0 and 100, with no additional commentary or text. This prompt allows for a more granular evaluation of the model's performance, especially in terms of accuracy and relevance.

These prompts are critical for evaluating different facets of model performance, ensuring that the model’s responses are both accurate and relevant.


\clearpage

\section{Full Related Work}

\subsection{Facial Video Analysis}
Face video analysis is a long standing problem towards high-level human understanding which involves various tasks, including attribute perception~\cite{zheng2020survey,cai2023marlin}, expression analysis~\cite{ben2021video,sun2023mae}, and emotion understanding~\cite{zhang2023weakly,wang2015video}.
Advanced models can be divided into those for attribution perception tasks~\cite{zhao2025humanomni,li2025faceinsight} and for high-level emotion understanding tasks~\cite{zhang2023weakly,wang2015video,ben2021video,lian2023mer}.
Various face perception models leverages strong backbone power for constructing multi-task framework~\cite{sun2024face}.
Pioneering methods are developed to solve pre-defined tasks, while MLLM-based method are proposed to enhance their zero-shot capacity.
%
Going forward to high-level emotion understanding~\cite{zhang2024masked,zhao2021emotion,lian2025dmer}, recent methods embrace MLLM for their strong zero-shot perception capacity~\cite{sun2024face,zhang-etal-2024-visual,li2024facial,Etesam2024ContextualER}. 
EMO~\cite{xing2024emo} firstly incorporates facial priors, including facial embeddings, landmarks, and age-gender-race attributes, through a mlp-based connnector to improve emotion understanding.
EmotionLLaMA~\cite{cheng2024emotionllama} introduces an emotion dataset including 28K coarse-grained and 4K fine-grained annotated datasets.
OmniEmotion~\cite{yang2025omni} proposes to explicitly integrate facial and audio modeling for emotion recognition. 
FacialDynamic~\cite{zhao2025facial} construct a existing largest human-labelled dataset with 5K samples.
ExpLLM~\cite{lan2025expllm} recently explore to leverage chain-of-thought strategy to empower LLM with the reasoning capability.
However, existing approaches are often constrained to a limited set of emotion categories or rely on static attribution perception.
To advance cognitive human emotion understanding, we propose a fine-grained emotion-centric model empowered by dynamic attribution perception and emotion reasoning.

\subsection{Video Extension in MLLM}
VideoLLMs~\cite{lin2024video,maaz2023video,zhang2024video,li2023videochat,maaz2024videogpt+} have gained significant attention in recent years by leveraging existing pre-trained foundational models, particularly powerful Large Language Models (LLMs), to enhance support for video inputs and outputs~\cite{cheng2024videollama,hong2024cogvlm2,zohar2024apollo,xu2024pllava}.
The key components of VideoLLMs include:
1) a video encoder responsible for encoding inputs from different modalities into feature representations that the model can understand, \textit{e.g.}, ViT~\cite{dosovitskiy2021an}, CLIP~\cite{radford2021learning};
2) an input projector to align encoded spatiotemporal features from video with textual feature space in LLM.
Input projectors can be implemented by linear projection or compressed-based projection such as Q-Former~\cite{li2023blip} or P-Former~\cite{jian2024bootstrapping};
and 3) LLM Backbone based on pre-trained models like GPT~\cite{achiam2023gpt} or LLaMA~\cite{touvron2023llama}, which processes representations from different modalities and performs semantic understanding.

\subsection{Reasoning Model in MLLM}
With the blossom of a series of recent models such as DeepSeek-R1 and OpenAI o-series~\cite{openai2025o3-mini,guo2025deepseek}, various works probe into integrating MLLMs with reasoning capacity~\cite{ahn-etal-2024-large}.
Multimodal chain-of-thought (MCoT) prompting~\cite{li2025imagine,thawakar2025llamav} offers a step-by-step reasoning trajectory when MLLM faces hard questions including detail grounding~\cite{wu2024v}, agent planing~\cite{li2025imagine}, etc.
Specifically, MCoT aims to tackle the question through several solving steps and a reasoning chain, enabling the generation of more effective results for complex problems step-by-step~\cite{xiang2025towards,qwen2024qvq,zhang2024improve}.
Recent works have demonstrated that CoT prompting substantially improves the MLLM’s capability on reasoning tasks.
For instance, LLaVA-CoT~\cite{xu2024llava} prompts MLLMs reasoning steps into the summary, caption, reasoning, and conclusion stages and proposes a stage-level beam search strategy to further enhance reasoning capacity.
LLaVA-Reasoner~\cite{zhang2024improve} pioneers the use of forced Chain-of-Thoughts, establishing a new direction for structured prompting techniques.
In this paper, we propose affective cues-based rationale tree as intermediate bridge to meet the gap between abstract emotion and basic attribute.

\clearpage

\section{Visualization}

In this section, we present visualization samples generated by our \model\ model, as well as those from our constructed \dataset\ dataset.
For more comprehensive visualization results, please refer to the video demos provided in the attached zip file.

\subsection{Results from our~\model~model}
\label{sec:vis_model}
\begin{figure}[ht]
  \centering
\includegraphics[width=0.85\linewidth]{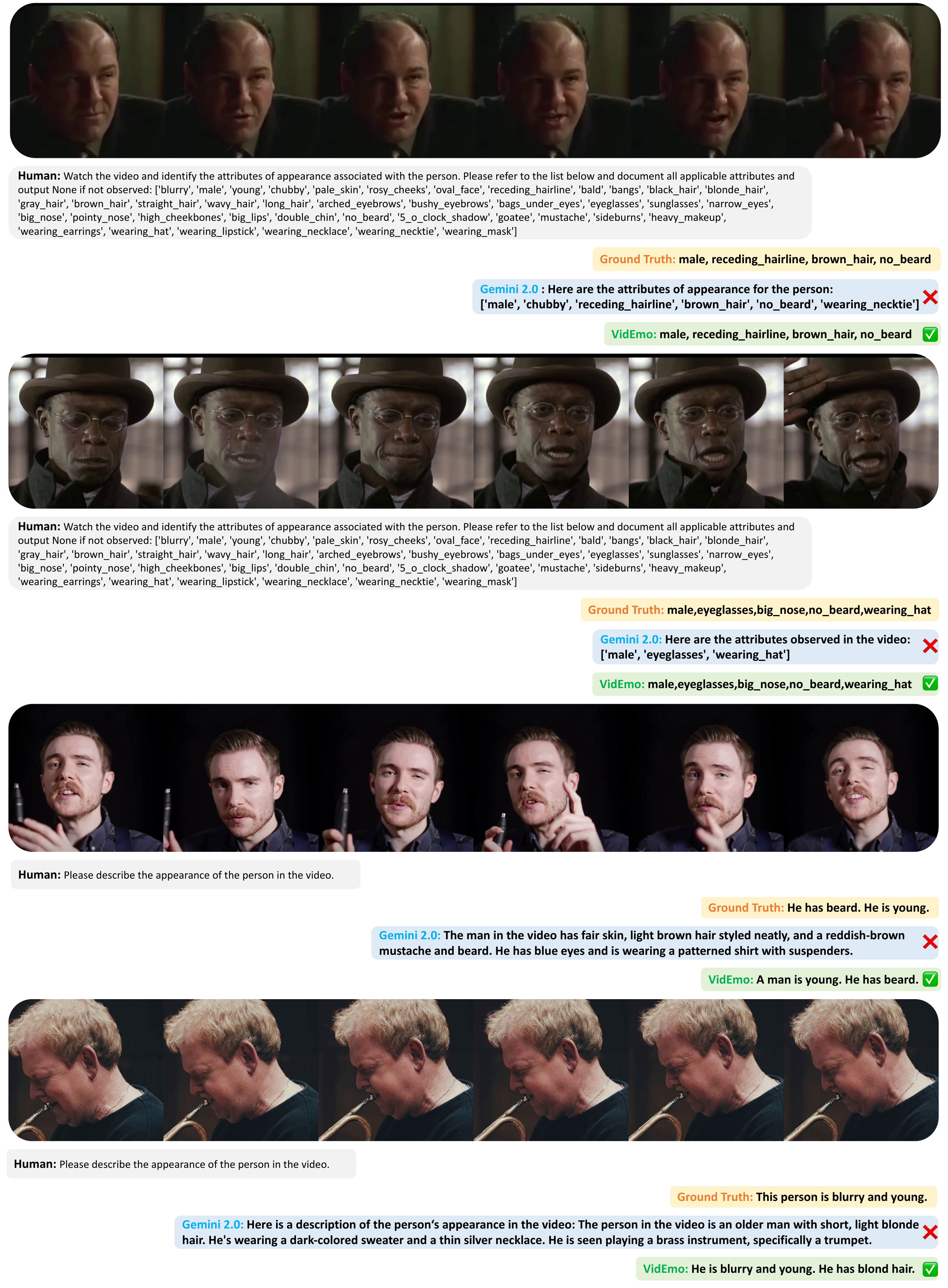}
  \caption{Visualization comparison results for appearance recognition and appearance caption. }
  \label{fig:appearance}
\end{figure}

\begin{figure}[ht]
  \centering
\includegraphics[width=0.85\linewidth]{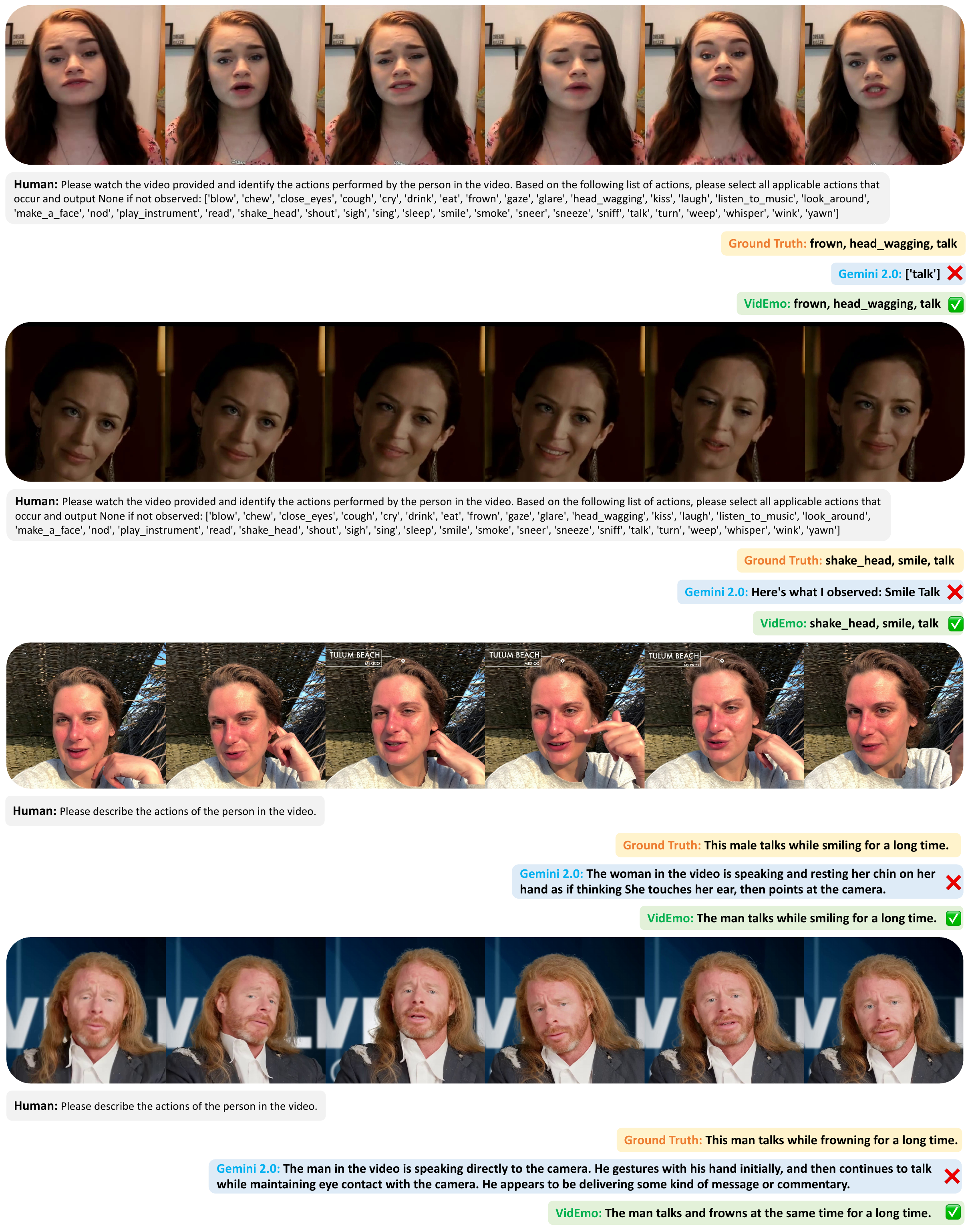}
  \caption{Visualization comparison results for action recognition and action caption. }
  \label{fig:action}
\end{figure}

\begin{figure}[ht]
  \centering
\includegraphics[width=0.85\linewidth]{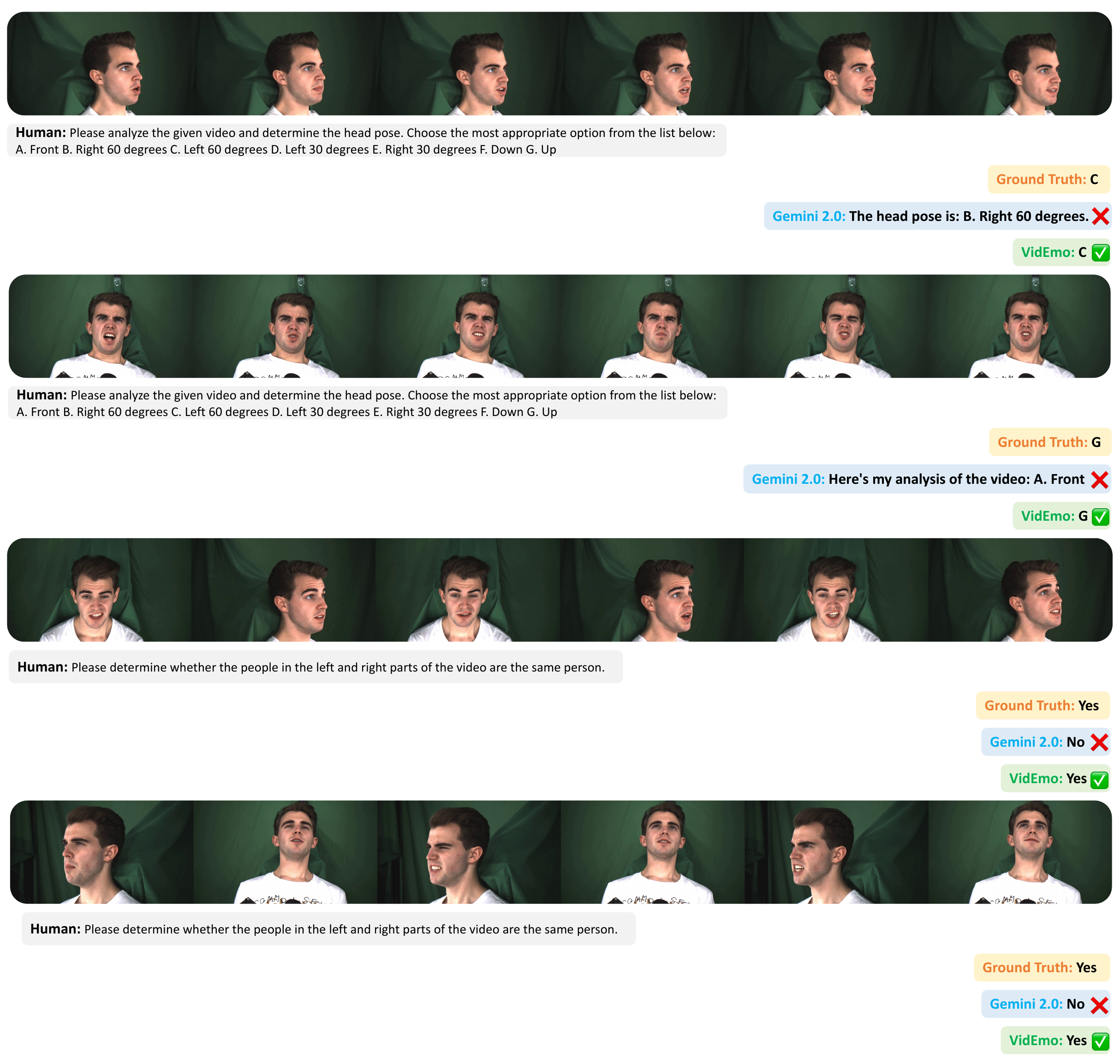}
  \caption{Visualization comparison results for head pose estimation and identity verification. }
  \label{fig:head}
\end{figure}

\begin{figure}[ht]
  \centering
\includegraphics[width=0.85\linewidth]{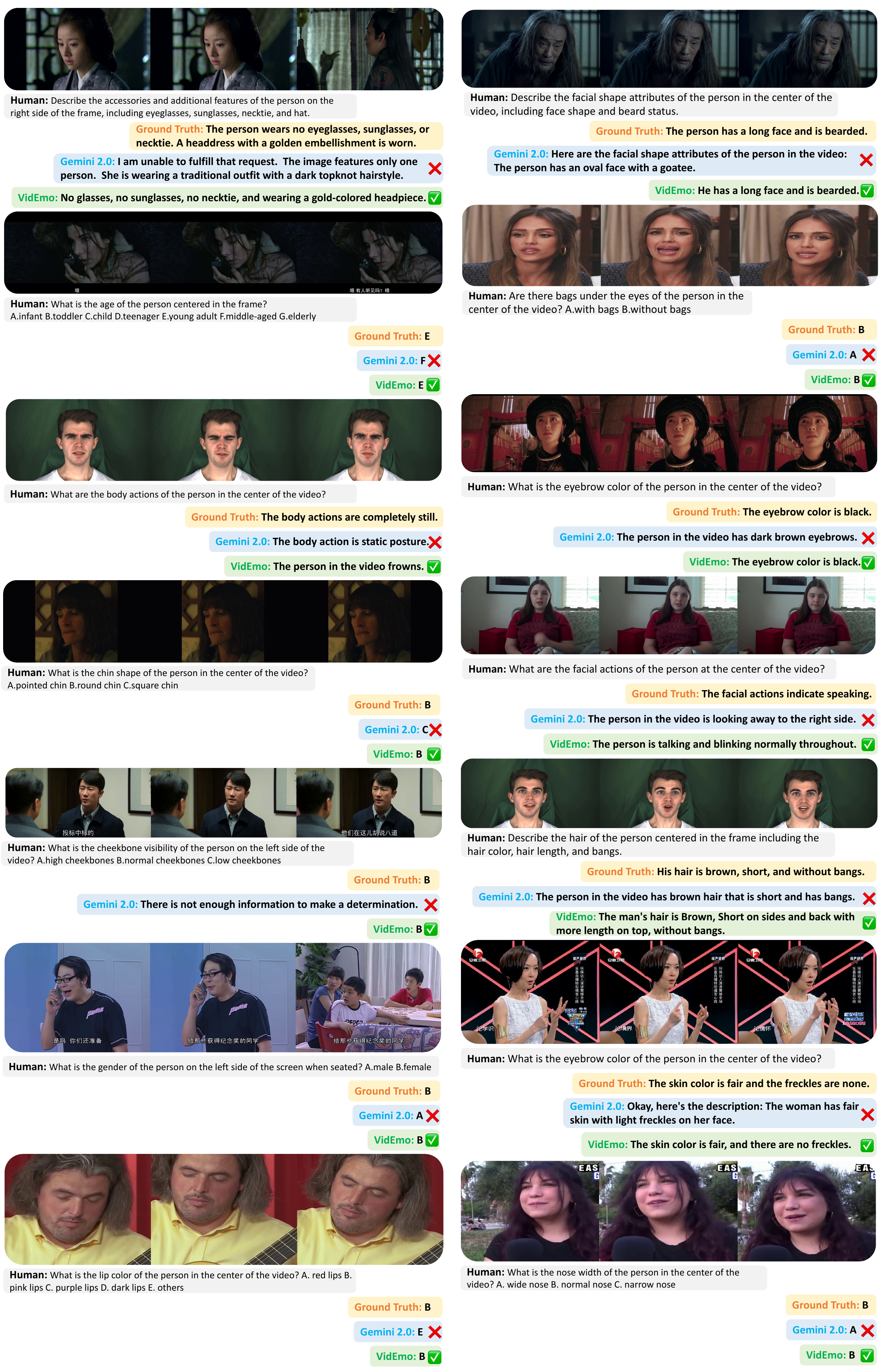}
  \caption{Visualization comparison results for open attribute perception. }
  \label{fig:QA}
\end{figure}

\textbf{Attribute Perception.}
We show the visualization comparison between our~\model and cutting-edge milestone Gemini 2.0 for appearance recognition and appearance caption in Figure~\ref{fig:appearance}, action recognition and action caption in Figure~\ref{fig:action}, head pose recognition and human identity recognition in Figure~\ref{fig:head}, open attribute perception in Figure~\ref{fig:QA}.

\begin{figure}[ht]
  \centering
\includegraphics[width=0.85\linewidth]{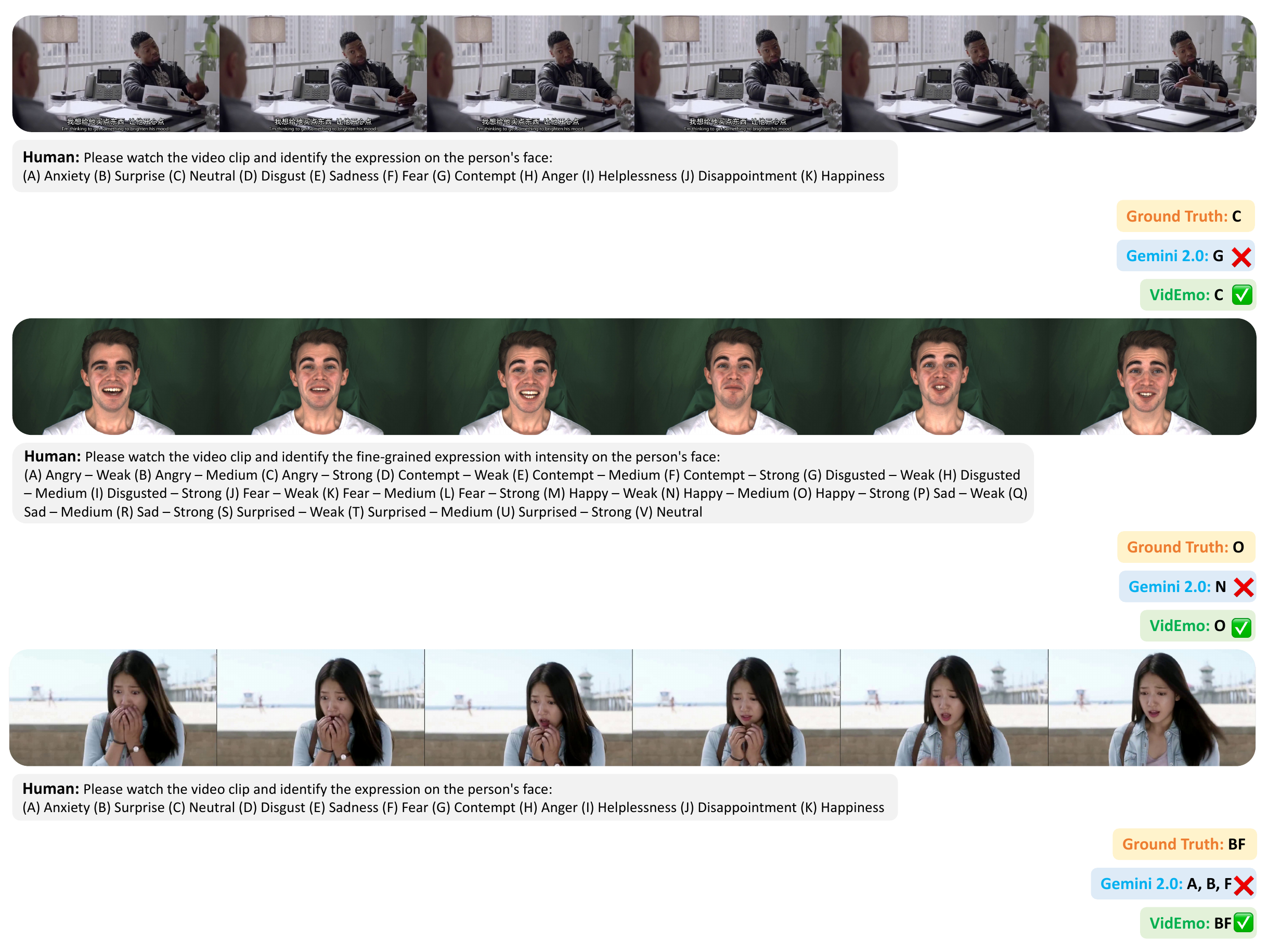}
  \caption{Visualization comparison results for single-label emotion recognition, multi-label emotion recognition and fine-grained emotion recognition. }
  \label{fig:er}
\end{figure}

\begin{figure}[ht]
  \centering
\includegraphics[width=0.85\linewidth]{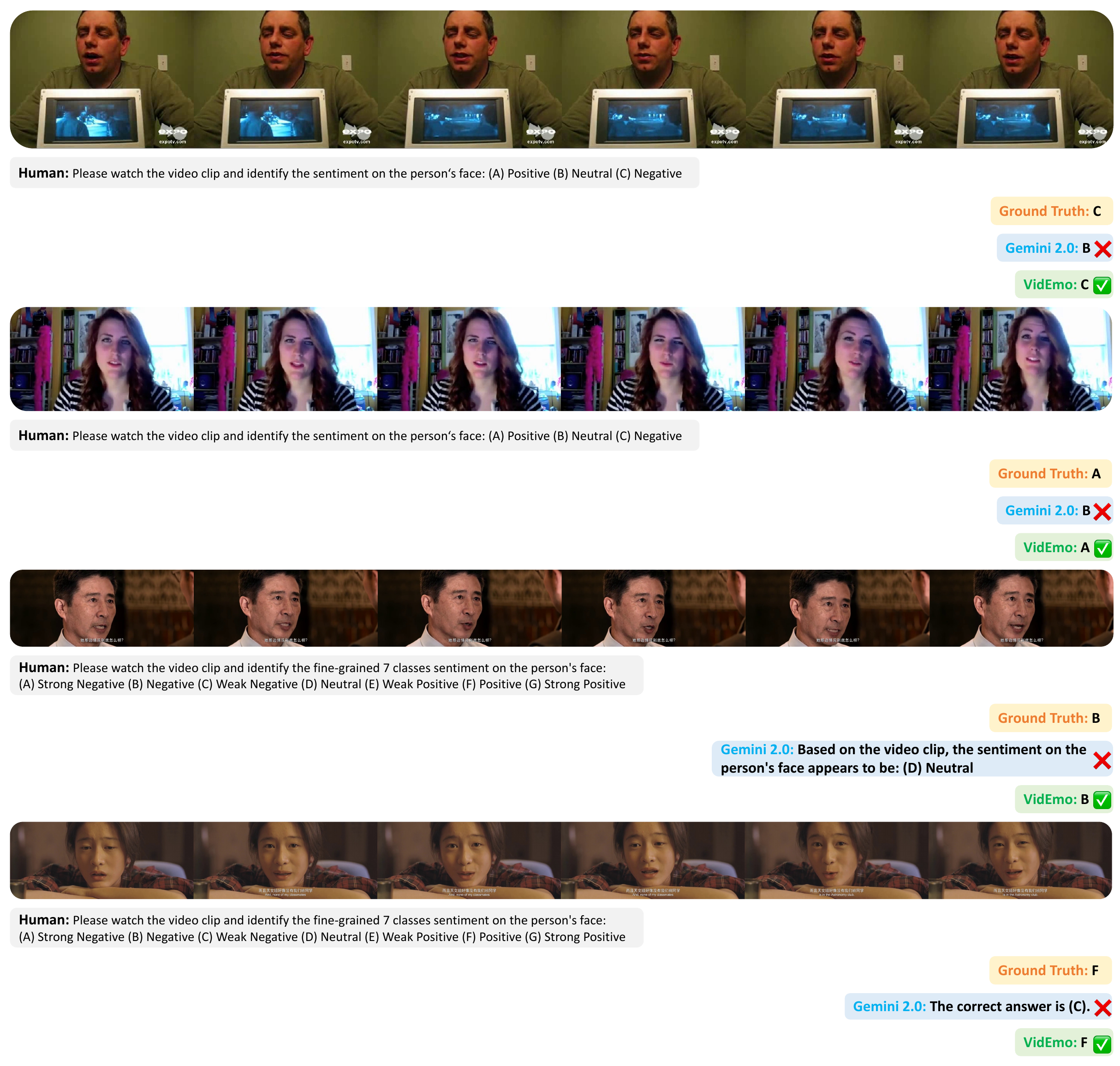}
  \caption{Visualization comparison results for micro-expression detection and action unit detection. }
  \label{fig:sr}
\end{figure}

\begin{figure}[ht]
  \centering
\includegraphics[width=0.85\linewidth]{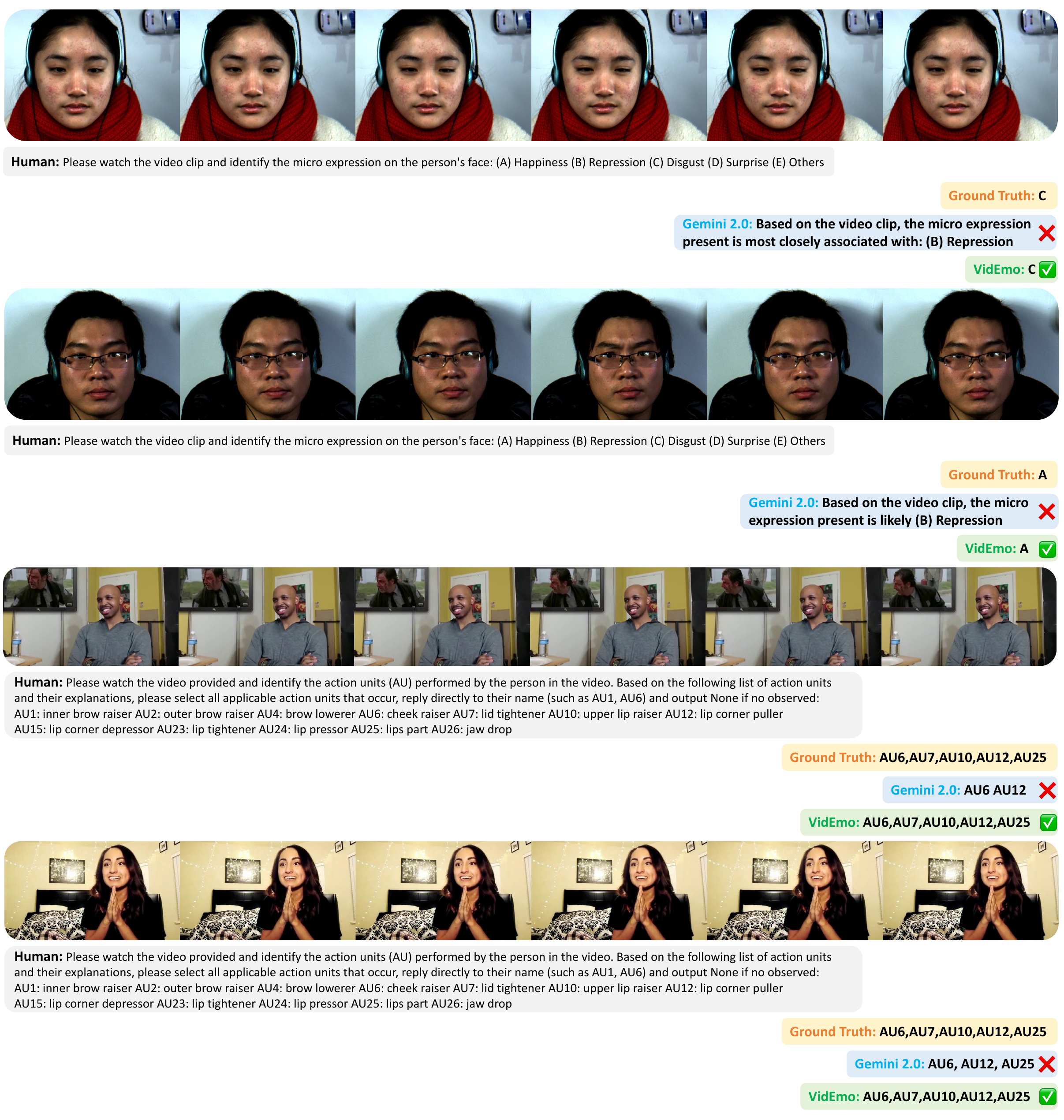}
  \caption{Visualization comparison results for single-label sentiment recognition and fine-grained sentiment recognition. }
  \label{fig:cues}
\end{figure}

\begin{figure}[ht]
  \centering
\includegraphics[width=0.85\linewidth]{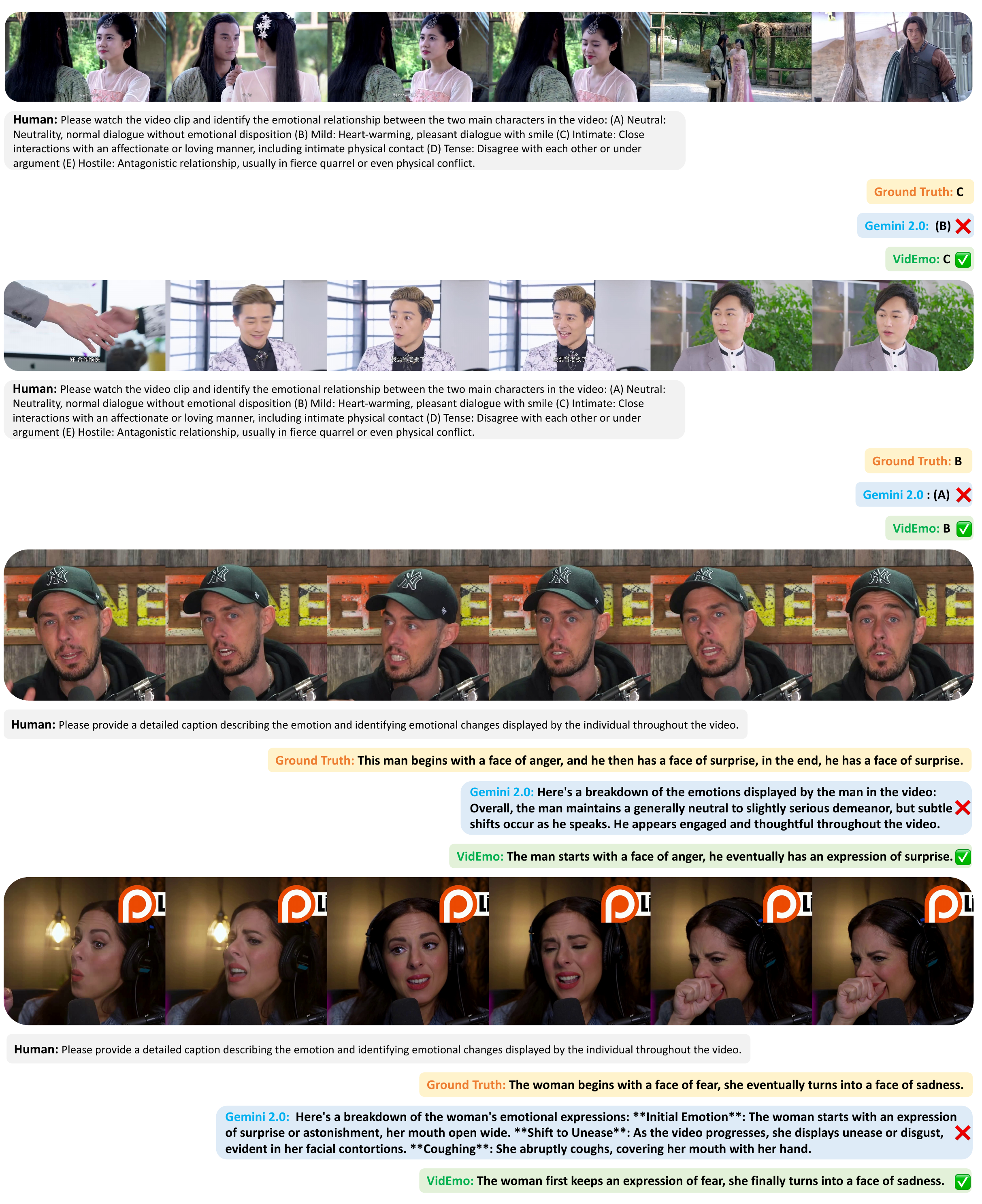}
  \caption{Visualization comparison results for conversation reasoning and emotion caption. }
  \label{fig:complex}
\end{figure}

\textbf{Expression Analysis.}
We show the visualization comparison between our~\model and cutting-edge milestone Gemini 2.0 for single-label emotion recognition, fine-grained emotion recognition and multi-label emotion recognition in Figure~\ref{fig:er}, single-label sentiment recognition and fine-grained sentiment recognition in Figure~\ref{fig:sr}, micro-expression detection and action unit detection in Figure~\ref{fig:cues}, conversation reasoning and emotion caption in Figure~\ref{fig:complex}.

\begin{figure}[ht]
  \centering
\includegraphics[width=0.85\linewidth]{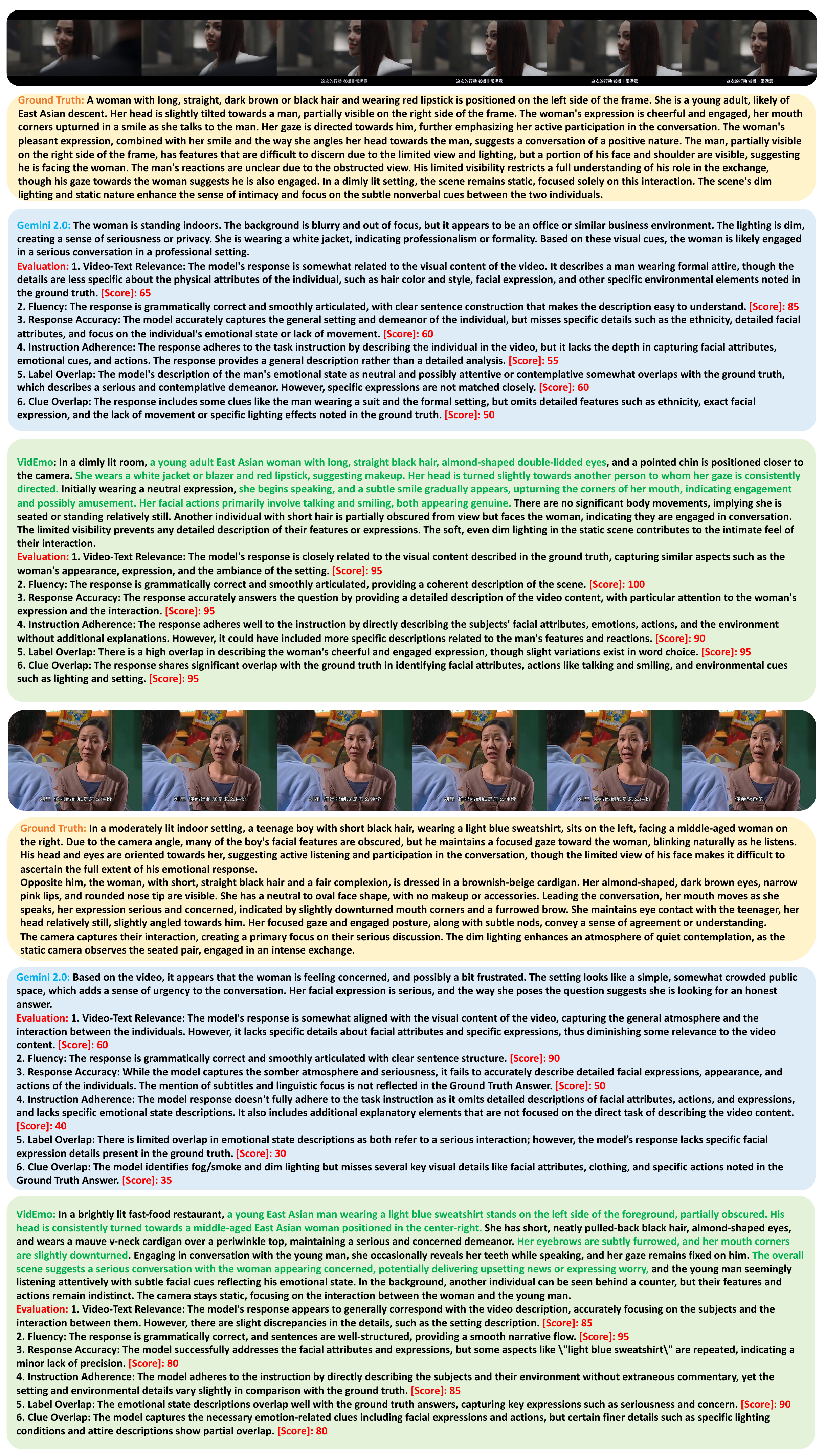}
  \caption{Visualization comparison results for fine-grained emotion caption. We achieve competitive performance with Gemini 2.0 on six different metrics.}
  \label{fig:caption_comp}
\end{figure}

\textbf{Emotion Understanding.}
We show the visualization comparison between our~\model and cutting-edge milestone Gemini 2.0 for fine-grained video caption in Figure~\ref{fig:caption_comp}

\subsection{Samples from our~\dataset~Dataset}

\begin{figure}[ht]
  \centering
\includegraphics[width=1.0\linewidth]{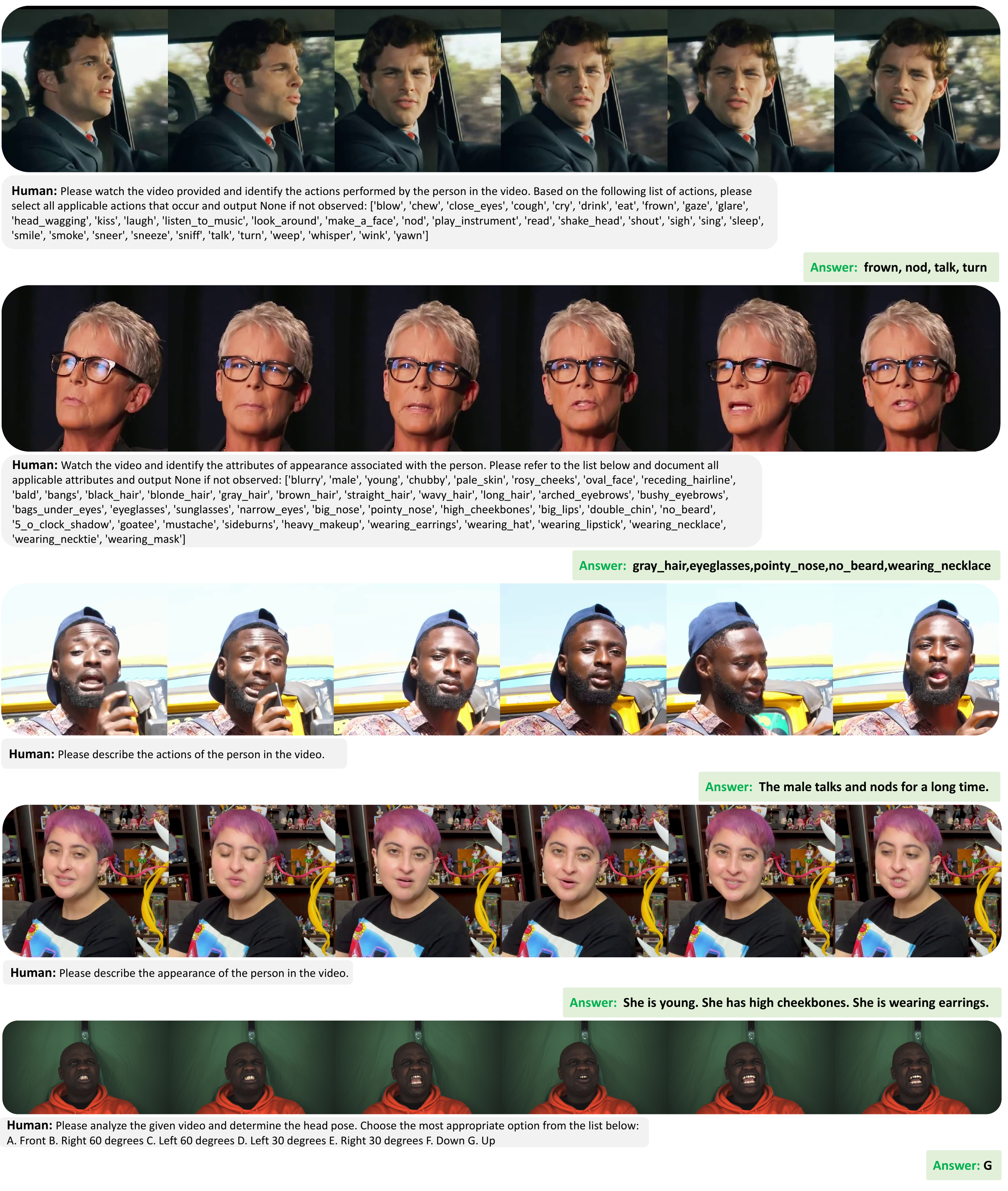}
  \caption{Visualization samples for attribute perception in classification-type tasks.}
  \label{fig:datasets_open_attri}
\end{figure}

\begin{figure}[ht]
  \centering
\includegraphics[width=1.0\linewidth]{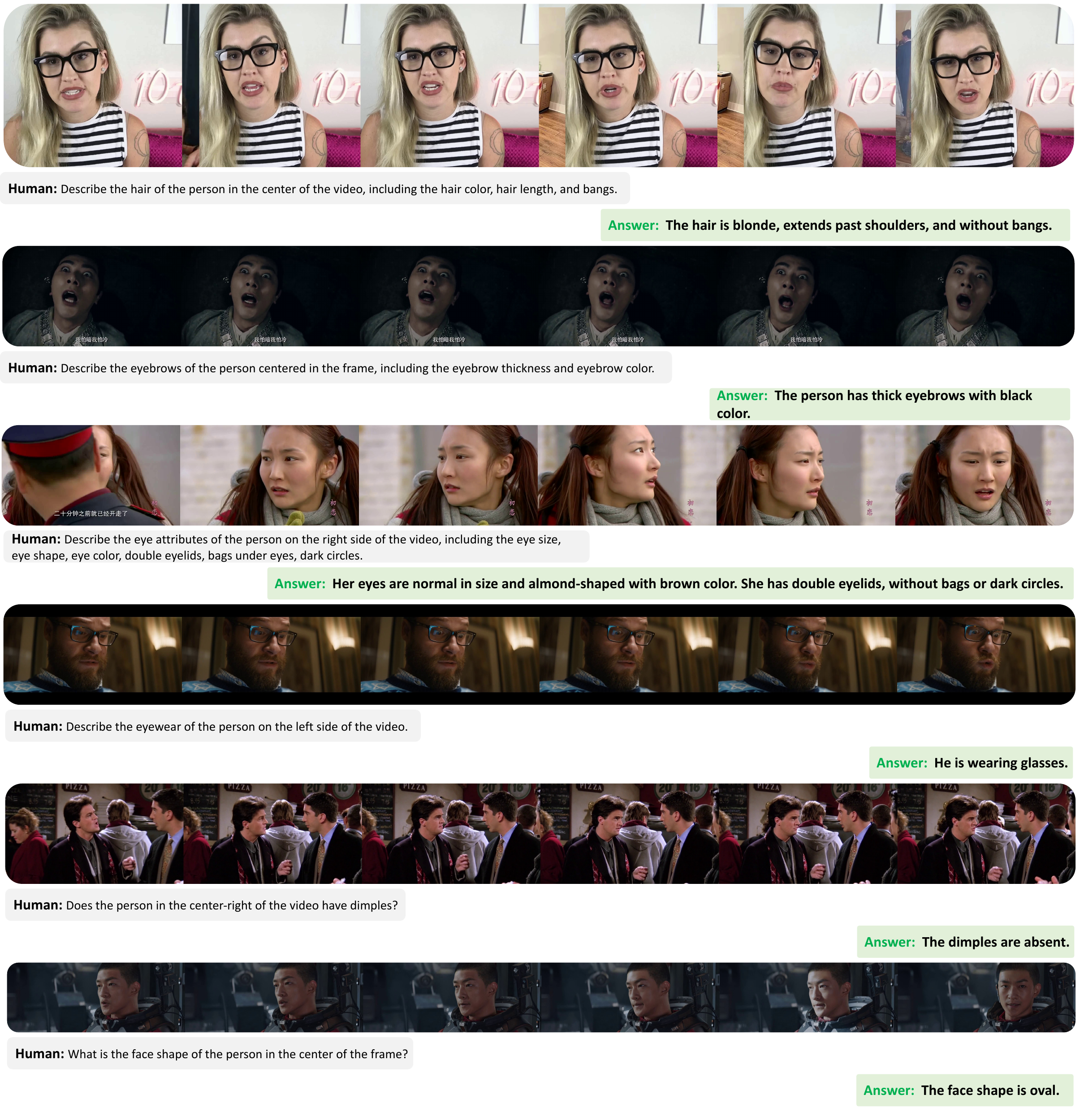}
  \caption{Visualization samples for attribute caption in caption-type tasks.}
  \label{fig:datasets_qa}
\end{figure}

\textbf{Attribute Perception.}
As shown in Figure~\ref{fig:datasets_open_attri}, we provide samples from public face attribute datasets, and we convert their annotations into question-answer pairs. 
In addition, we visualized the labeled face attribute samples in Figure~\ref{fig:datasets_qa}.

\begin{figure}[ht]
  \centering
\includegraphics[width=1.0\linewidth]{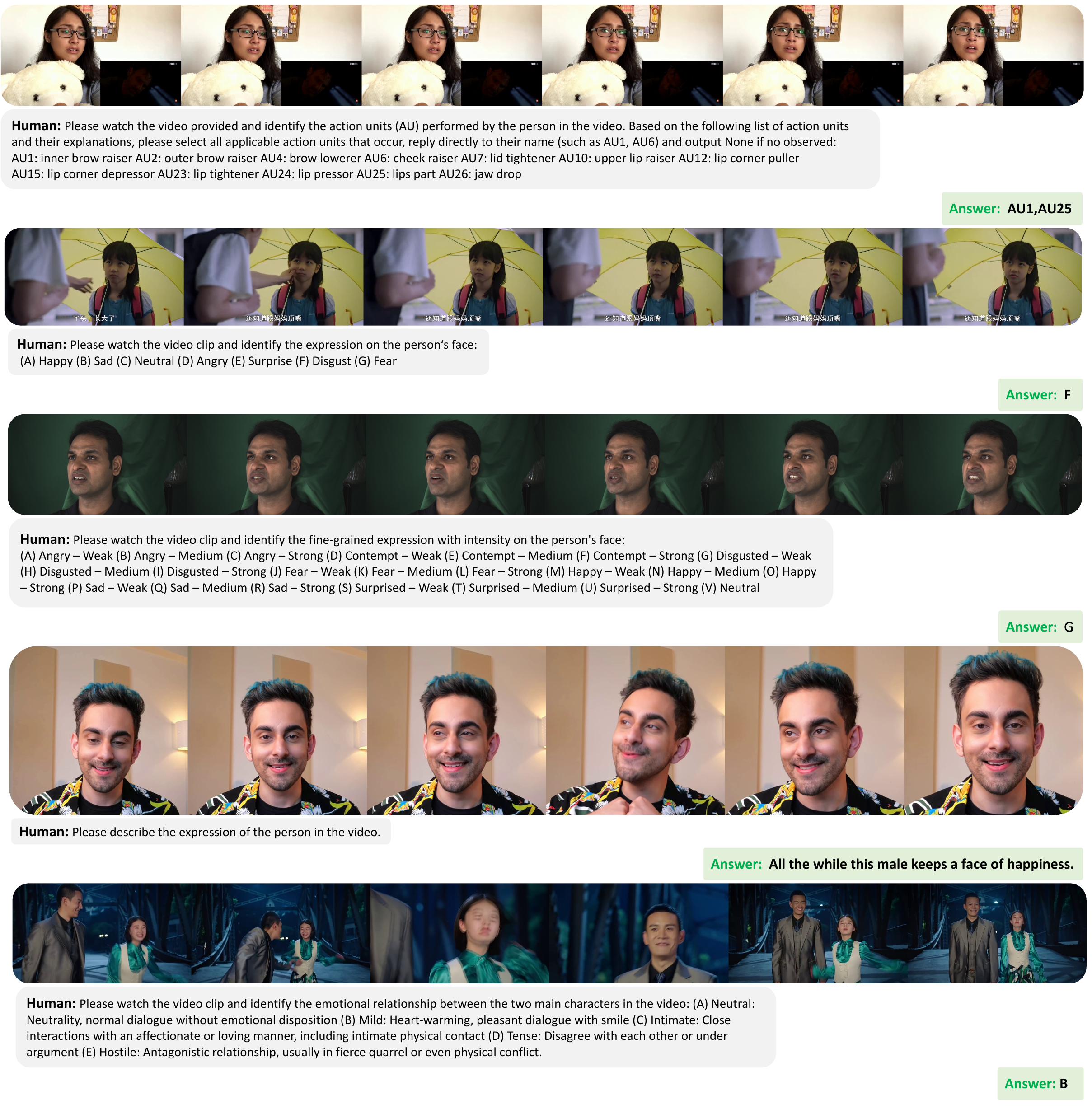}
  \caption{Visualization samples for expression analysis.}
  \label{fig:datasets_open_emotion}
\end{figure}

\begin{figure}[ht]
  \centering
\includegraphics[width=1.0\linewidth]{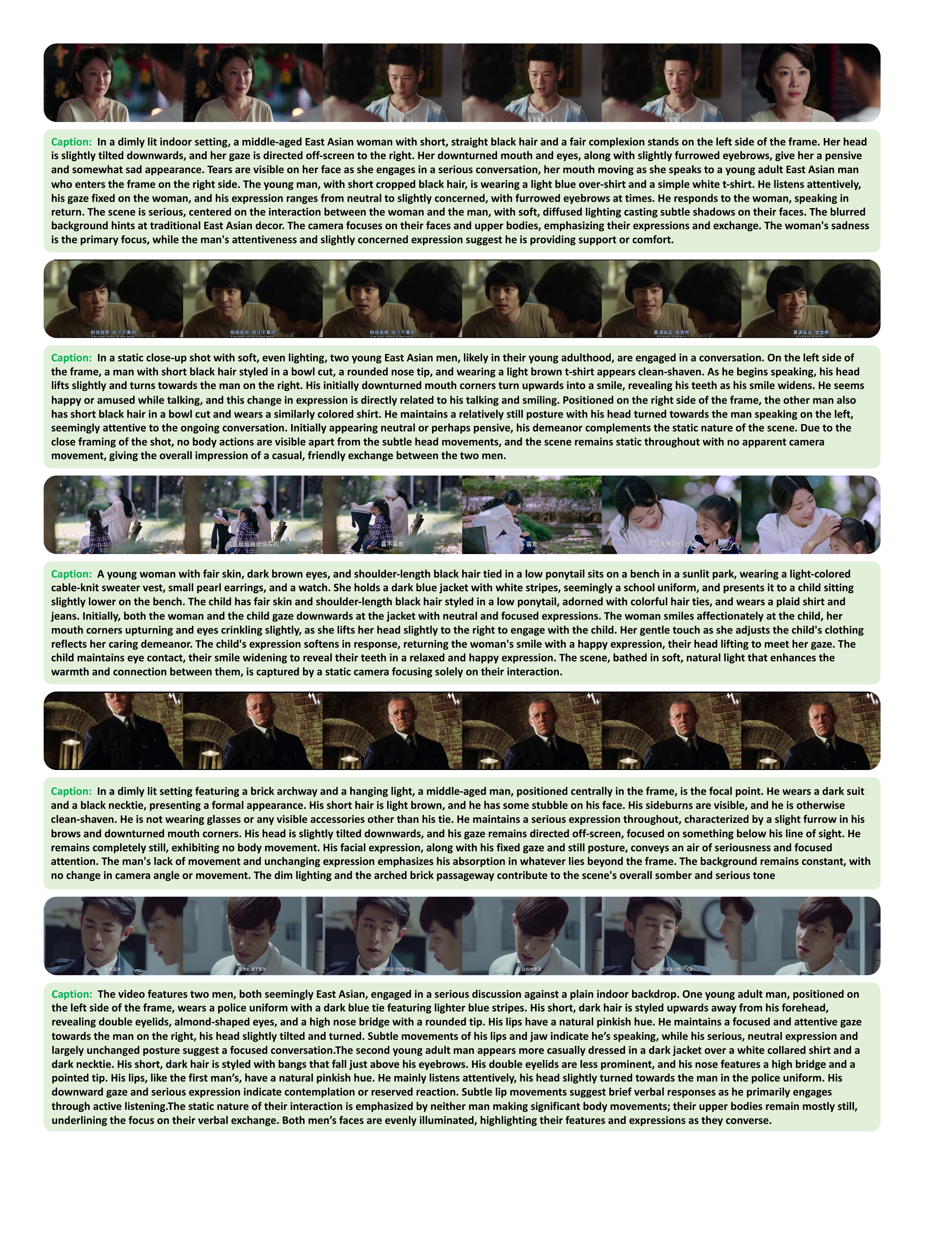}
  \caption{Visualization samples for emotion understanding in fine-grained caption task.}
  \label{fig:datasets_caption}
\end{figure}

\textbf{Expression Analysis.}
As shown in Fig~\ref{fig:datasets_open_emotion} and Fig~\ref{fig:datasets_caption}, we provide samples from public emotion recognition datasets and our fine-grained caption dataset.

\begin{figure}[ht]
  \centering
\includegraphics[width=1.0\linewidth]{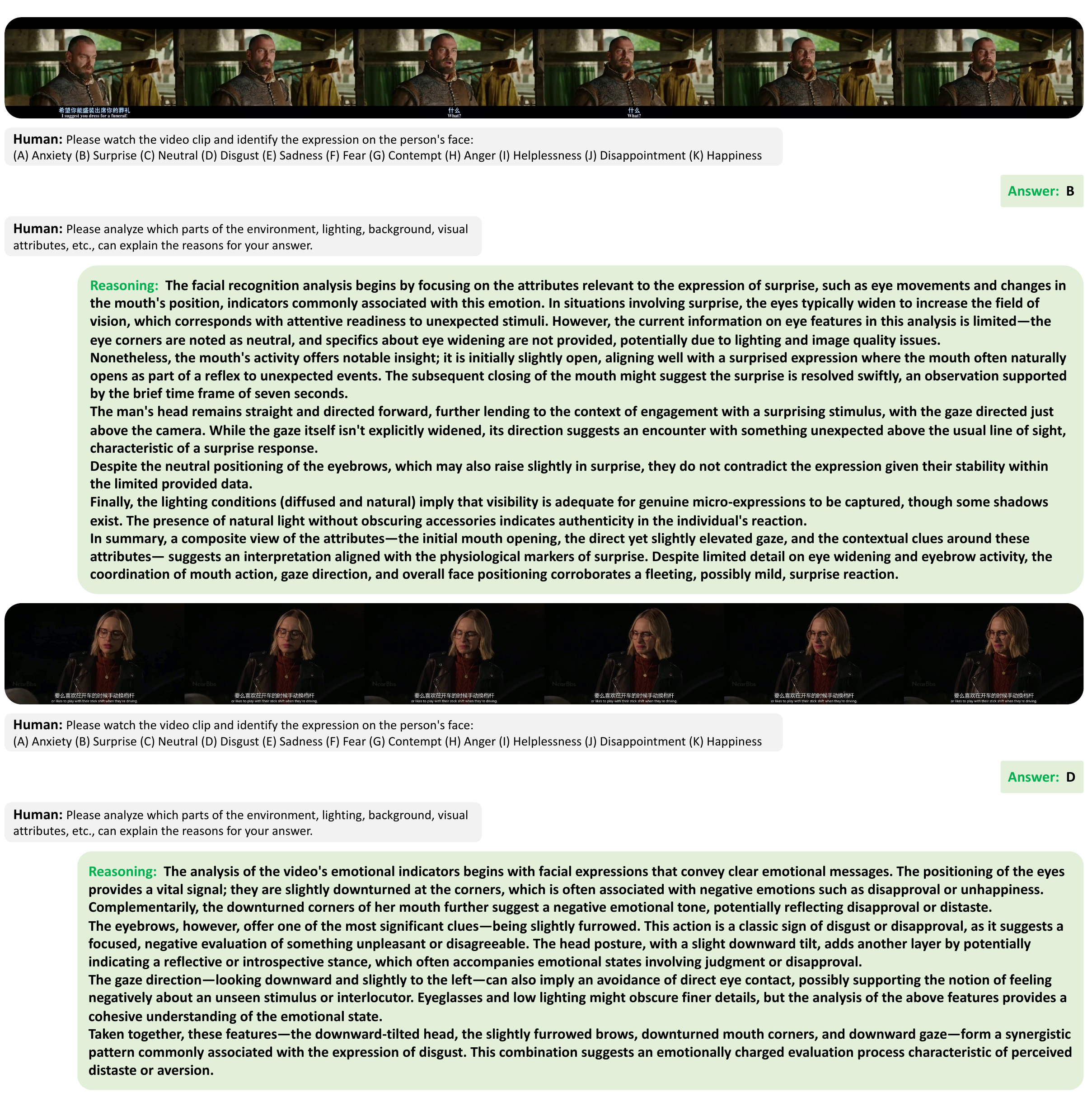}
  \caption{Visualization samples for emotion understanding in rationale analysis task.}
  \label{fig:datasets_rational}
\end{figure}

\textbf{Emotion Understanding.}
As shown in Fig~\ref{fig:datasets_rational}, we provide samples for emotion reasoning from the provided dataset.

\begin{figure}[ht]
  \centering
\includegraphics[width=1.0\linewidth]{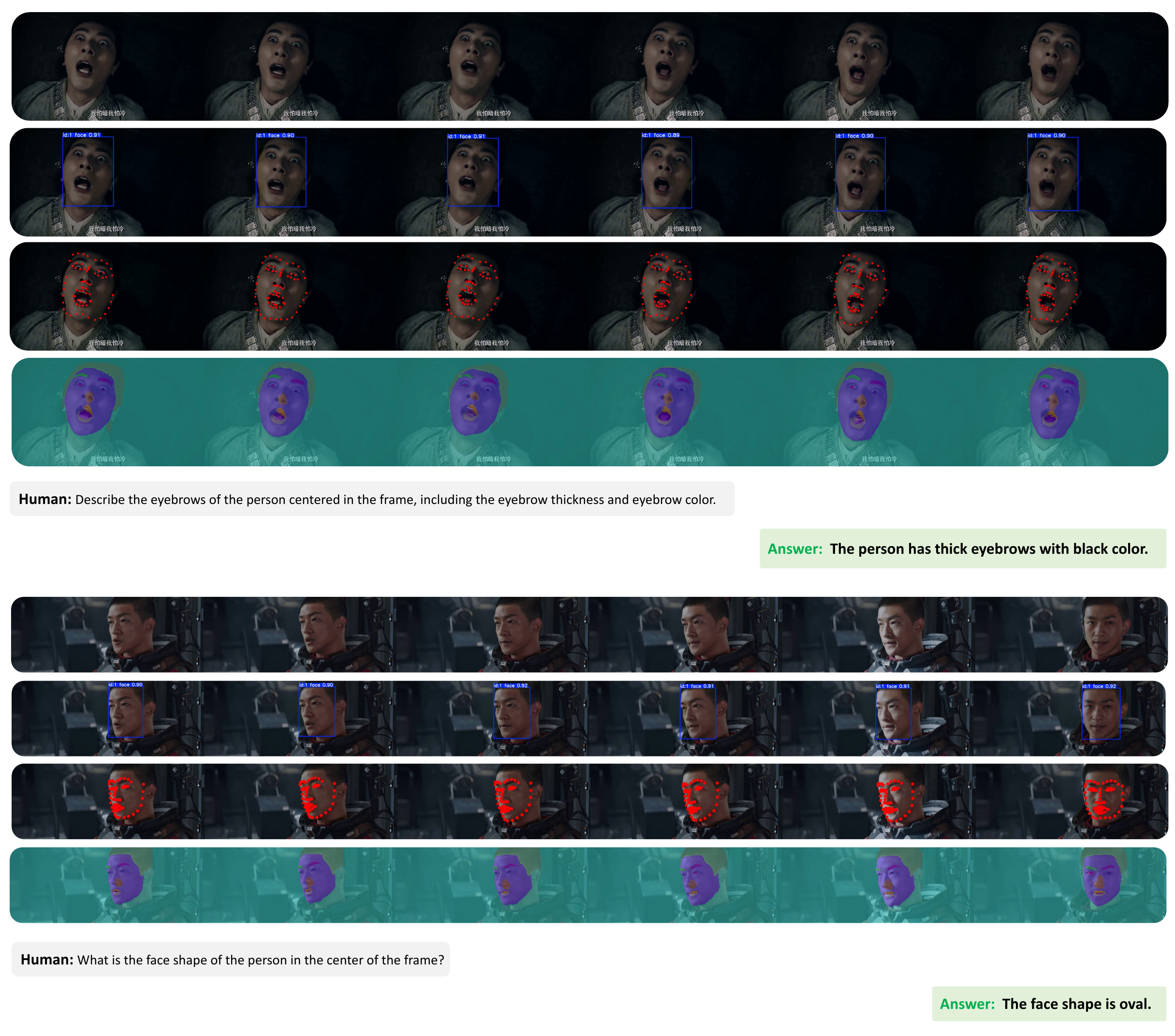}
  \caption{Visualization samples of meta labels of face boxes, face landmarks and parsing masks.}
  \label{fig:datasets_meta}
\end{figure}

\textbf{Meta Labels of Face Landmarks and Parsing Masks.}
As shown in Fig~\ref{fig:datasets_meta}, we provide meta labels of face boxes, face landmarks and parsing masks.


\clearpage 

\end{document}